%% file: arxiv.tex
\documentclass[10pt,twocolumn,letterpaper]{article}

\usepackage[pagenumbers]{cvpr}
\usepackage[T1,OT1]{fontenc}
\input{preamble}

\definecolor{cvprblue}{rgb}{0.21,0.49,0.74}
\usepackage[breaklinks,colorlinks,allcolors=cvprblue]{hyperref}

\title{Image Diffusion Models Are Training-Free Pixel Art Generators If Constrained}
\author{{\fontencoding{T1}\selectfont Arthus de Senneville} \\ Independent
\and {\fontencoding{T1}\selectfont Adhemar de Senneville} \\ Independent}
\hypersetup{%
    pdftitle={Image Diffusion Models Are Training-Free Pixel Art Generators If Constrained},
    pdfauthor={Arthus de Senneville and Adhémar de Senneville}
}

\begin{document}

\maketitle

\input{sec/0_abstract}
\input{sec/1_intro}

\input{sec/2_Related_work}
\input{sec/3_BACKGROUND}

\input{sec/4_METHOD}

\input{sec/5_RESULTS}

\input{sec/6_CONCLUSION}

\input{sec/7_ACKNOWLEDGMENTS}
\pdfbookmark[1]{LLM Usage Statement}{preprint-llm-usage}
\input{sec/8_LLM_USAGE}

{\fontsize{9pt}{10.5pt}\selectfont
    \setlength{\bibsep}{0pt}
    \pdfbookmark[1]{References}{preprint-references}
    \bibliographystyle{ieeenat_fullname}
    \bibliography{main}
}

\input{sec/X_suppl}

\end{document}

%% file: preamble.tex
\usepackage{booktabs}
\usepackage{multirow}
\usepackage{graphicx}
\usepackage{pifont}

\newcommand{\cmark}{\ding{51}}
\newcommand{\xmark}{\ding{55}}

\newcommand{\PAPx}{\operatorname{PAP}}

\AtBeginDocument{%
  \everydisplay\expandafter{\the\everydisplay
    \setlength{\abovedisplayskip}{4pt plus 1pt minus 1pt}%
    \setlength{\belowdisplayskip}{4pt plus 1pt minus 1pt}%
    \setlength{\abovedisplayshortskip}{0pt plus 1pt}%
    \setlength{\belowdisplayshortskip}{3pt plus 1pt minus 1pt}%
  }%
}

\makeatletter
\renewcommand{\paragraph}{\@startsection{paragraph}{4}{\z@}%
  {1ex}
  {-1em}
  {\normalfont\normalsize\bfseries}}
\makeatother

%% file: sec/0_abstract.tex
\begin{abstract}
Pixel art uses a coarse grid and a limited color palette to depict recognizable objects with very few pixels. 
It remains widely used in video games, digital icons, embroidery, and bead mosaics. General-purpose diffusion models are, on the surface, good pixel art generators.
They can reproduce pixel art conventions such as crisp outlines, simplified shading, and dithering.
Also, unlike specialized models trained on narrow pixel art datasets, they can follow a wide range of prompt instructions.
However, upon closer inspection, the generated pixel art fails to meet the basic requirements for grid alignment, grid size, and color palette that downstream applications depend on. 
In this work, we propose PAPify, a training-free method that constrains a diffusion model at inference time to meet these requirements. 
PAPify projects intermediate image predictions onto a fixed grid and palette during denoising, allowing the model to adapt the image to those constraints.
Despite being simple, fast, and computationally efficient, PAPify achieves state-of-the-art performance compared with previous specialized pixelization and generation methods.
\end{abstract}


%% file: sec/1_intro.tex
\section{Introduction}
\label{sec:intro}

\begin{figure}[!t]
    \centering
    \includegraphics[width=\columnwidth]{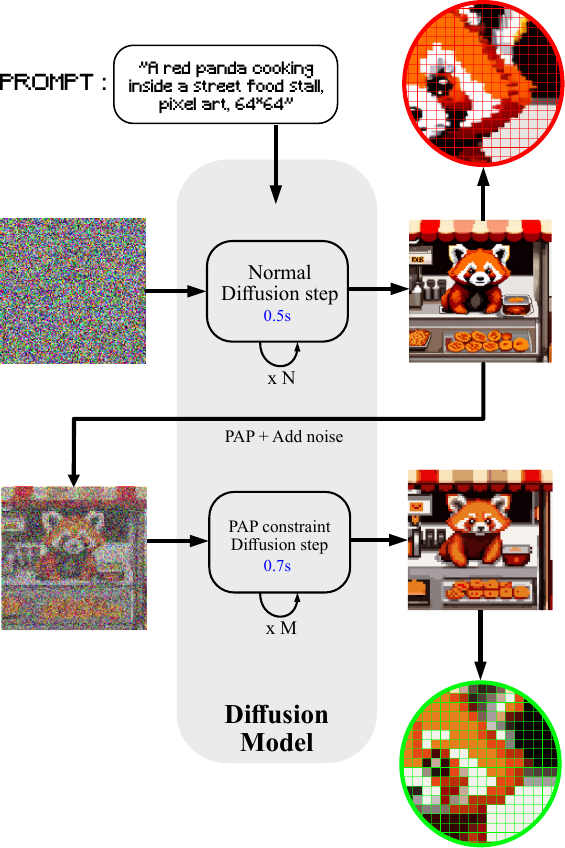}
    \caption{Overview of PAPify. A first diffusion pass generates a reference image, which is projected onto the target pixel art space and partially noised. A second, PAP-constrained pass transforms the image into pixel art while preserving its semantic content.}
    \label{fig:method}
\end{figure}

\begin{figure*}[!t]
    \centering
    \includegraphics[width=\textwidth]{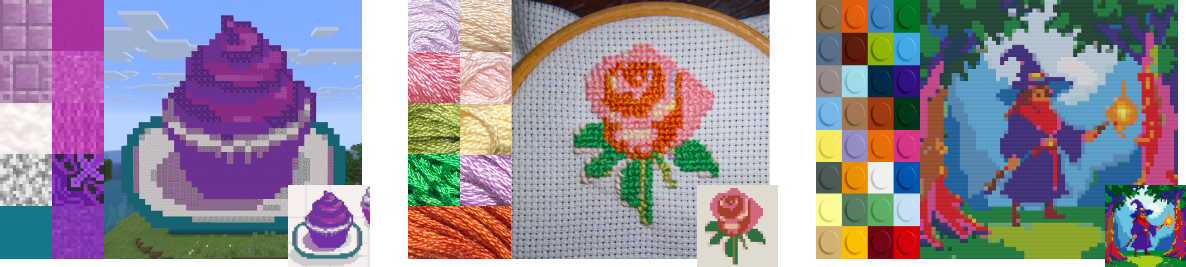}
    \vspace{-4mm}
    \caption{Examples of pixel art applications, including game assets, embroidery, and interlocking-brick mosaics.}
    \label{fig:example}
    \vspace{-3mm}
\end{figure*}

Recent advances in text-to-image generation have transformed visual-content production and increasingly affect creative industries such as 
illustration, animation, and video games. 
Among these domains, pixel art remains particularly important. Originally developed under severe hardware and memory limitations, 
pixel art represents images using a low-resolution spatial grid and a restricted set of colors. 
Despite these historical origins, it remains widely used in modern games, digital assets, icons, and physical crafts such as embroidery, fuse-bead mosaics, and interlocking-brick designs, where each stitch or material unit can naturally correspond to one pixel \cite{binninger2024sdpixl,gerstner2012pixelated}.

However, pixel art cannot be reduced to ordinary images displayed at low resolution. It relies on deliberate artistic conventions governing edge sharpness, contrast, color transitions, dithering, silhouettes, and pixel clusters \cite{gerstner2012pixelated,wu2022sprites,Matusovic2023Depixelization}. Consequently, a simple Pixel Art Projection (PAP), consisting of spatial downsampling onto a fixed grid followed by color quantization to a predefined palette, may satisfy constraints while still producing blurred contours, noisy gradients, and other artifacts inconsistent with authentic pixel art
\cite{gerstner2012pixelated,wu2022sprites}.


Current text-to-image diffusion models can imitate the visual appearance of pixel art through prompting, yet often fail to satisfy its discrete constraints. 
Generated pixels may vary in size, contain internal texture, or fall outside the target palette, as shown in \cref{fig:method}. 
As discussed above, applying PAP only after generation does not ensure visually coherent pixel art.

A direct alternative is to train a generative model exclusively on pixel art datasets.
However, their limited scale \cite{binninger2024sdpixl} and unbalanced style distributions \cite{chen2025gametilenet} can limit generalization to unusual scenes \cite{coutinho2022generating,li2025pixdiff}. Adapting such models to new resolutions, pixel block sizes, or palettes may also require additional data or retraining \cite{binninger2024sdpixl}.

Lastly, some lines of work adapt general-purpose diffusion models to produce images that respect both the discrete constraints and aesthetic conventions of pixel art \cite{binninger2024sdpixl, jain2023vectorfusion}. However, these methods often rely on gradient computation, making image generation too slow for consumer use.


In this work, we introduce PAPify, a training-free method that uses pretrained diffusion models to generate pixel art under explicit spatial and color constraints. 
Rather than applying PAP only after generation, PAPify integrates it directly into the denoising process by repeatedly projecting intermediate clean-image predictions. 
This allows the diffusion model to progressively adapt its semantic structure and local composition to the target pixel art representation.
PAPify requires no backpropagation, enabling faster generation than gradient-based approaches.


Our contributions are:
\begin{itemize}
    \item We introduce PAPify for text-to-pixel art generation and image pixelization with user-defined target block size resolutions and color palettes.
    \item We propose a flow shift adaptation based on the target pixel block size, motivated by the lower effective noise level perceived by the model after block averaging.
    \item We benchmark PAPify against prior pixelization and generation methods. In our setup, it achieves approximately $140\times$ faster generation while using half the GPU memory of the previous state of the art.
\end{itemize}



%% file: sec/2_Related_work.tex
\section{Related Work}
\label{sec:related_work}

Existing approaches to pixel art synthesis can be divided into two main categories: pixelization methods, which transform existing images into pixel art, and generative methods, which synthesize new images from prompts. We review both directions, focusing on their ability to maintain aesthetic quality and visual coherence while satisfying spatial and color constraints.

\subsection{Pixelization}

\paragraph{Classical Methods.}
Traditional pixelization relies on spatial downsampling and color quantization. Gerstner et al.~\cite{gerstner2012pixelated} introduced Pixelated Image Abstraction (PIA), jointly optimizing spatial regions and palette selection to preserve perceptually important structures. Shang and Wong~\cite{shang2021automatic} subsequently proposed an automatic portrait pixelization method that preserves facial features and sharp contours. Although these methods offer control over resolution and color, their reliance on the original image structure limits their ability to reinterpret shapes and composition toward authentic pixel art aesthetics, unlike our generative approach.

\paragraph{Neural Methods.}
Learning-based pixelization methods use neural networks to translate conventional images into pixel art. Deep Unsupervised Pixelization (DUP)~\cite{han2018deep} learns this transformation without paired training examples, while Kuang et al.~\cite{kuang2021pixel} adopt a CycleGAN-based approach. Make Your Own Sprites (MYOS)~\cite{wu2022sprites} explicitly models aliasing and pixel-cell structure to produce sharper and more controllable results. While these methods learn pixel art characteristics, their generalization is limited by their training datasets, unlike our approach, which benefits from the diversity of general-purpose diffusion models.

\subsection{Pixel Art Generation}

\paragraph{Specialized Generative Models.}
Dedicated pixel art generators learn the target style through training or fine-tuning. PixDiff-PIG~\cite{li2025pixdiff} trains a compact diffusion model using palette-aware representations and text guidance. Pixel Art XL~\cite{pixelartxl} adapts SDXL~\cite{podell2024sdxl} through LoRA~\cite{hu2022lora}, while PixelLab AI~\cite{pixellab} provides specialized pixel art generation and editing tools. Such approaches learn pixel art conventions directly, but their capabilities depend on the diversity of their training data.

\paragraph{General-Purpose Generative Models.}
Large text-to-image models, including Stable Diffusion 3.5~\cite{stabilityai2024sd35}, FLUX.2 [klein] 4B (FLUX)~\cite{blackforestlabs2026flux2klein4b}, and GPT-Image-2.5 Sunburst (GPTImage)~\cite{openai2026gptimage25sunburst}, offer broad semantic coverage and stylistic flexibility.
Although they can imitate pixel art through prompting, they do not consistently respect the requested pixel art resolution, grid alignment, or color palette.
Additional projection is therefore needed to satisfy these constraints, often at the expense of aesthetic quality.

\paragraph{Constrained Diffusion Generation.}
VectorFusion~\cite{jain2023vectorfusion} demonstrated how pretrained diffusion priors can guide the optimization of structured visual representations through Score Distillation Sampling (SDS). 
Building on this principle, SD-$\pi$XL~\cite{binninger2024sdpixl} generates pixel art by optimizing a discrete, palette-constrained representation using latent score distillation. 
It supports arbitrary resolutions and palettes without pixel art-specific training.
However, both methods rely on costly gradient-based optimization at inference time. For example, SD-$\pi$XL requires approximately 1.5--2.5 hours per image on an NVIDIA RTX 4090 GPU~\cite{binninger2024sdpixl}.


\begin{table}[t]
\centering
\caption{Comparison of pixel art synthesis methods.}
\label{tab:method_comparison}
\small
\setlength{\tabcolsep}{2pt}
\renewcommand{\arraystretch}{0.95}
\setlength{\aboverulesep}{1.5pt}
\setlength{\belowrulesep}{1.5pt}

\begin{tabular*}{\columnwidth}{@{\extracolsep{\fill}}clccccr@{}}
\toprule
& \textbf{Method}
& \textbf{Hard}
& \textbf{Resolution}
& \textbf{Text}
& \textbf{T-free}
& \textbf{Time} \\
\midrule

\multirow{6}{*}{\rotatebox{90}{Pixelization}}
& PIA~\cite{gerstner2012pixelated} & \cmark & \cmark & \xmark & \cmark & 6\,s \\
& APIP~\cite{shang2021automatic} & \xmark & \cmark & \xmark & \cmark & 1\,s \\
& AOP~\cite{lei2024art} & \xmark & \cmark & \xmark & \cmark & 0.3\,s \\
& DUP~\cite{han2018deep} & \xmark & \xmark & \xmark & \xmark & 0.6\,s \\
& MYOS~\cite{wu2022sprites} & \xmark & \cmark & \xmark & \xmark & 0.2\,s \\
& \textbf{PAPify (ours)} & \cmark & \cmark & \cmark & \cmark & \textbf{17\,s} \\

\midrule

\multirow{6}{*}{\rotatebox{90}{Generation}}
& SD 3.5~\cite{stabilityai2024sd35} & \xmark & \xmark & \cmark & \cmark & 30\,s \\
& FLUX~\cite{blackforestlabs2026flux2klein4b} & \xmark & \xmark & \cmark & \cmark & 6\,s$^\dagger$ \\
& GPTImage~\cite{openai2026gptimage25sunburst} & \xmark & \xmark & \cmark & \cmark & 21\,s$^\dagger$ \\
& VectorFusion~\cite{jain2023vectorfusion} & \xmark & \cmark & \cmark & \cmark & 25\,min \\
& SD-$\pi$XL~\cite{binninger2024sdpixl} & \cmark & \cmark & \cmark & \cmark & 1.5\,h \\
& \textbf{PAPify (ours)} & \cmark & \cmark & \cmark & \cmark & \textbf{40\,s} \\

\bottomrule
\end{tabular*}

\vspace{1pt}
\parbox{\columnwidth}{%
\footnotesize
Hard: hard constraints; Text: text conditioning; T-free: training-free.
$\dagger$: API time; full experimental details, including hardware specifications, are provided in the supplementary material.
}
\end{table}


\vspace{0.2cm}

%% file: sec/3_BACKGROUND.tex
\section{Background}


\subsection{Flow Matching and Sampling}
\label{sec:flow_matching}

PAPify operates at the sampling level and can be adapted to models using Rectified Flow or its variants. 
We therefore review in this section this Flow Matching formulation~\cite{lipman2023flow,liu2023flow}, along with flow shifted sampling~\cite{esser2024scaling} and Euler integration.

\paragraph{Theory.}
We denote pixel-space images by $x$ and latent representations by $z$, with $\mathcal{E}$ and $\mathcal{D}$ denoting the VAE encoder and decoder, respectively.
Given an image $x_0 \sim p_{\mathrm{data}}$, its encoding $z_0=\mathcal{E}(x_0)$, and Gaussian noise
$\epsilon \sim \mathcal{N}(0,I)$, Rectified Flow defines the linear probability path
\begin{equation}
    z_t = (1-t)z_0 + t\epsilon,
    \qquad t\in[0,1],
\end{equation}
with associated velocity
\begin{equation}
    v_t = \frac{d z_t}{dt} = \epsilon - z_0.
\end{equation}
A neural network $v_\theta(z_t,t)$ is trained to approximate this velocity by minimizing
\begin{equation}
    \mathcal{L}_{\mathrm{FM}}
    =
    \mathbb{E}_{t,z_0,\epsilon}
    \left[
        \left\|
        v_\theta(z_t,t)-v_t
        \right\|_2^2
    \right].
\end{equation}

\paragraph{Integration.}
At inference, generation starts from $z_1\sim\mathcal{N}(0,I)$ and follows the learned ODE $\frac{dz_t}{dt}=v_\theta(z_t,t)$ backward from noise to data. Following Stable Diffusion 3.5~\cite{stabilityai2024sd35,esser2024scaling}, we use the flow shifted schedule $\sigma(t;m)=\frac{mt}{1+(m-1)t}$, where $m>1$ allocates more steps to high noise levels. Given timesteps $1=t_0>\cdots>t_N=0$, we define $\sigma_i=\sigma(t_i;m)$. We integrate this trajectory using the explicit Euler method, requiring one model evaluation per step:
\begin{equation}
    z_{\sigma_{i+1}}
    =
    z_{\sigma_i}
    +
    (\sigma_{i+1}-\sigma_i)
    v_\theta(z_{\sigma_i},\sigma_i).
    \label{eq:euler_update}
\end{equation}

\begin{figure*}[!t]
    \centering
    \includegraphics[width=\textwidth]{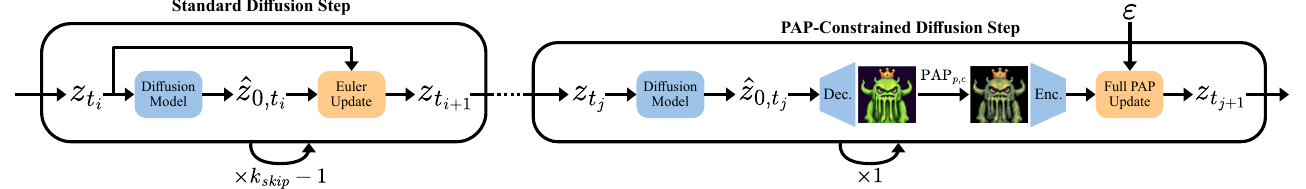}
    \caption{Overview of PAPify's second inference stage. Periodic PAP updates project clean-image predictions onto the target grid and palette, guiding subsequent diffusion steps to transform the image into coherent pixel art.}

    \label{fig:steps}
\end{figure*}


\subsection{Pixel Art Projection (PAP)}
\label{sec:PAP}

We define \emph{Pixel Art Projection} (PAP) as the transformation of an arbitrary image into a pixel art representation satisfying prescribed spatial and color constraints. As discussed in Sec.~\ref{sec:related_work}, various pixelization methods can perform this transformation.
In this work, we adopt the conventional PAP approach, denoted $\PAPx_{p,c}$, which performs spatial downsampling with pixel block width $p$ followed by nearest-color quantization under color constraint $c$.
For convenience, we keep the PAP output at the same spatial dimensions as the input image throughout this paper.

%% file: sec/4_METHOD.tex
\section{PAPify Method}

In this section, we introduce a training-free method that guides a pretrained text-to-image model toward pixel art generation. Our method requires only a text prompt and the desired pixel art characteristics.
It comprises two components. 
First, we introduce the PAPify pipeline, a two-stage inference procedure: it generates a reference image, then refines it into pixel art. This refinement relies on the introduced Full PAP Update, a modification of Euler sampling (\cref{fig:steps}).
Second, we adapt the flow shift parameter to the target pixel size, modifying the denoising schedule so that the sampling dynamics remain compatible with larger pixel blocks.


\subsection{PAPify Pipeline}

Our inference procedure is composed of two complementary stages. 
The first inference constructs a semantically rich scene, while the second transforms this representation into pixel art through repeated PAP-constrained diffusion steps.

\paragraph{First Stage.}
Given a text prompt, a resolution parameter $p$, and a color palette $c$, we first perform standard text-to-image inference to obtain a high-quality reference image $x^{\mathrm{ref}}_0$ whose semantic content, global composition, object geometry, and fine-scale details will guide pixel art synthesis in the second stage.
The prompt does not need to mention pixel art, as our method independently steers generation toward this style during refinement.
%
Then, we project the decoded reference image as $x^{\mathrm{PAP}}_0=\PAPx_{p,c}(x^{\mathrm{ref}}_0)$ and encode it as $z^{\mathrm{PAP}}_0=\mathcal{E}(x^{\mathrm{PAP}}_0)$.

\paragraph{Second Stage (Pixelization).}
We initialize the second stage by partially noising the PAP-projected latent \(z^{\mathrm{PAP}}_0\) to level \(\sigma_s\). This preserves the reference image’s content while allowing the model to adapt its appearance to pixel art conventions:
\begin{equation}
z_{\sigma_s}
=
(1-\sigma_s)z^{\mathrm{PAP}}_0
+ \sigma_s \epsilon,
\qquad
\epsilon \sim \mathcal{N}(0,I).
\end{equation}

During denoising, we apply PAP every $k_{skip}$ sampling steps. At each projection step, the model’s clean latent prediction $\hat{z}_{0,i}$ is decoded into image space, projected using PAP, and re-encoded into latent space:
\begin{equation}
\resizebox{0.9\columnwidth}{!}{$\displaystyle
    \hat{z}^{\mathrm{PAP}}_{0,i}
    =
    \mathcal{E}\!\left(
    \PAPx_{p,c}\!\left(
        \mathcal{D}(\hat{z}_{0,i})
    \right)
    \right),
    \qquad
    i \equiv 0 \pmod{k_{skip}}.
$}
\end{equation}

The resulting PAP-projected prediction is then incorporated
into the denoising trajectory through the Full PAP Update
introduced in Sec.~\ref{sec:pap_euler}. We refer to these
sampling steps as \textit{PAP-constrained steps}.
After denoising, we apply a final PAP to the decoded image to enforce the target grid and palette.

\paragraph{Structural Conditioning.}
This repeated projection explicitly exposes the generative model to the desired discrete pixel structure throughout denoising. As discussed in the flow shift analysis in Sec.~\ref{sec:pixel_flow_shift}, the spatial organization induced by PAP provides a strong structural signal from which the model can recognize the image as belonging to the pixel art domain.
\begin{figure*}[!t]
    \centering
    \includegraphics[width=\textwidth]{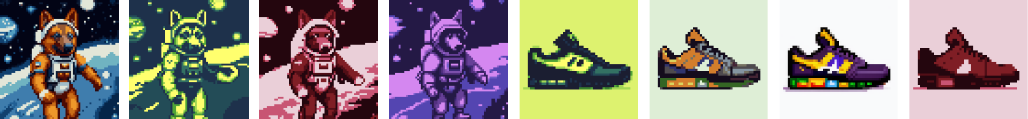}
    \vspace{-6mm}
    
    \caption{Pixel art generated by PAPify using different color palettes. PAPify adapts semantic representations and shading to the available colors, even with highly restricted palettes.}
    
    \label{fig:palette}
    \vspace{-3mm}
\end{figure*}
Consequently, once conditioned by the projected structure, the model can recover stylistic characteristics associated with pixel art.

This ability relies on the assumption that the underlying generative model has been exposed to pixel art imagery during training in order to guide generation toward the learned pixel art style.
Furthermore, we hypothesize that some pixel art training captions omit the style, explaining why PAPify does not require it in the prompt.


\subsection{Full PAP Update}
\label{sec:pap_euler}

Directly projecting the predicted clean latent in a standard Euler sampler during a PAP-constrained diffusion step has limited influence on the sampling trajectory, as the projected prediction contributes only partially to the next state.
To strengthen this influence, we introduce a \textit{Full PAP Update} that replaces the entire clean-latent component with its PAP-projected counterpart at selected sampling steps $i \equiv 0 \pmod{k_{skip}}$:
\begin{equation}
    z_{\sigma_{i+1}}
    =
    (1-\sigma_{i+1})\,\hat{z}^{\mathrm{PAP}}_{0,i}
    +
    \sigma_{i+1}\,\epsilon,
    \label{eq:full_pap}
\end{equation}
where $\hat{z}^{\mathrm{PAP}}_{0,i}$ is the PAP-projected clean latent and $\epsilon \sim \mathcal{N}(0,I)$ is the noise sampled to initialize the second stage, reused unchanged in every Full PAP Update.

Between PAP-constrained diffusion steps, sampling follows the standard Euler update in Eq.~\ref{eq:euler_update}. While we adopt Euler for simplicity, the same principle can be extended to higher-order samplers by modifying their predicted clean latents.


\subsection{Noise Scale Adaptation}
\label{sec:pixel_flow_shift}

In our method described above, we apply PAP during sampling to steer the diffusion process.
However, in this section, we show how this interferes with the denoising trajectory and how to mitigate the effect.
More specifically, we observe that PAP modifies the observed signal-to-noise amplitude ratio (SNAR).
For a noisy pixel-space image $x_\sigma=(1-\sigma)x_0+\sigma\epsilon$, the SNAR is
\begin{equation}
\mathrm{SNAR}(x_\sigma)
=\frac{\operatorname{Std}[(1-\sigma)x_0]}{\operatorname{Std}[\sigma\epsilon]}
=\frac{(1-\sigma)\operatorname{Std}[x_0]}{\sigma}.
\end{equation}
Here, $c=\varnothing$ disables color quantization, so $\PAPx_{p,\varnothing}$ is a linear spatial projection that averages pixel and noise values separately.
Averaging pixel values over a block has a small effect on their standard deviation,
$\operatorname{Std}[\PAPx_{p,\varnothing}(x_0)]\approx\operatorname{Std}[x_0]$.
In contrast, averaging \(p^2\) independent noise values in each block makes their fluctuations partially cancel, reducing the standard deviation by a factor of \(p\): \(\operatorname{Std}[\PAPx_{p,\varnothing}(\epsilon)]=1/p\).
This gives
\begin{equation}
\resizebox{0.9\columnwidth}{!}{$\displaystyle
\mathrm{SNAR}(\PAPx_{p,\varnothing}(x_\sigma))
=\frac{\operatorname{Std}[(1-\sigma)\PAPx_{p,\varnothing}(x_0)]}
      {\operatorname{Std}[\sigma \PAPx_{p,\varnothing}(\epsilon)]}
\approx p\,\mathrm{SNAR}(x_\sigma),
$}
\end{equation}
meaning that after projection, there is $p$ times less noise relative to the image. 
To counterbalance this effect, we adjust $\sigma$ according to the pixel size such that
\begin{equation}
\frac{1-\sigma_{p}}{\sigma_{p}}
=\frac{1}{p}\frac{1-\sigma}{\sigma}.
\end{equation}
This is equivalent to multiplying the flow shift by $p$.
To apply the same reasoning to latent variables $z$, we assume that applying PAP to the image produces approximately the same latent embedding as applying PAP to its latent embedding.


\subsubsection{Empirical Validation}
\label{sec:flowshift_analysis}
We conduct two experiments with a shared reconstruction protocol: we encode images of randomly colored $p\times p$ blocks and add noise at different levels.
At each timestep, SD3.5 Medium independently predicts a reconstruction using an empty prompt.
We compare the reconstructions visually and measure image-space MSE against the source images, averaged over ten images per configuration.

\begin{figure}[!t]
    \centering
    \includegraphics[
        width=1\columnwidth,
        keepaspectratio
    ]{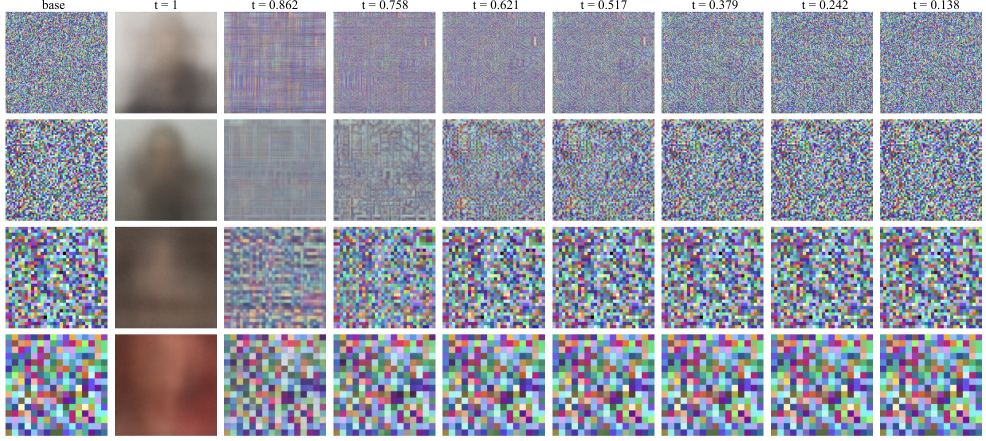}
    \caption{Reconstruction of random block-color images for different block sizes and noise levels, with $m=1$. The first column shows the VAE reconstruction of each source image; subsequent columns show independently predicted clean images decoded from noisy latents.}
    \label{fig:PAP_preserved}
\end{figure}

\paragraph{PAP across Image and Latent Spaces.}
Figure~\ref{fig:PAP_preserved} shows that blockwise patterns remain recognizable in the VAE reconstructions and decoded model predictions, particularly for larger blocks.
This provides qualitative support for the image--latent consistency assumption used in our theoretical analysis, suggesting that the spatial structure introduced by PAP is sufficiently preserved across these representations.
These preserved spatial patterns may also help the model recognize pixel art structure and guide its refinement during denoising.

\begin{figure}[!t]
    \centering
        \makebox[\columnwidth][c]{%
            \includegraphics[
                width=1.05\columnwidth,
                keepaspectratio
            ]{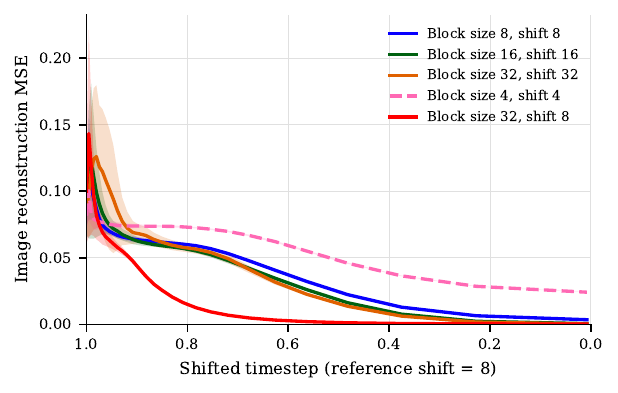}%
        }
    \caption{Image reconstruction MSE for different block sizes $p$ and flow shifts $m$, averaged over ten images per configuration. Proportional scaling ($m=p$) yields similar error curves for $p \geq 8$.}
    \label{fig:flow_shift_scaling}
\end{figure}

\paragraph{Flow Shift Scaling.}
In Fig.~\ref{fig:flow_shift_scaling}, the configurations $(p,m)=(8,8)$, $(16,16)$, and $(32,32)$ produce closely aligned error curves despite different target block size resolutions.
These results support multiplying the flow shift by $p$, as predicted in Sec.~\ref{sec:pixel_flow_shift}.
The scaling becomes less accurate for $(p,m)=(4,4)$ and smaller blocks.
We attribute this deviation to the VAE's limited ability to preserve fine spatial structures, as illustrated by the reconstruction artifacts for small blocks in the VAE baseline of Fig.~\ref{fig:PAP_preserved}.
This limitation motivates maintaining sufficiently large pixel blocks. Rather than decreasing the block size to generate higher-resolution pixel art, we increase the diffusion model's image resolution while keeping the block size fixed.
Empirically, we find that $p=8$ provides the best results.

%% file: sec/5_RESULTS.tex

\section{Experiments}
\label{sec:results}

\begin{figure*}[!t]
    \centering
    \includegraphics[width=\textwidth]{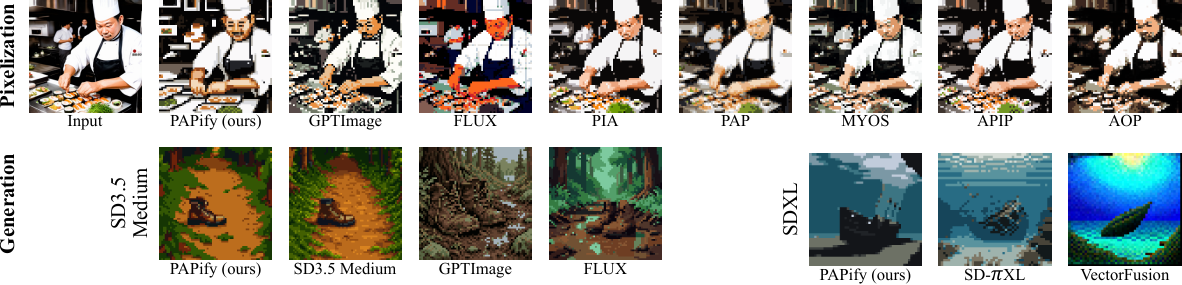}
    \vspace{-5mm}
    \caption{Qualitative comparison of PAPify with pixelization and generation baselines. Generation results include comparisons using both SD3.5 Medium and SDXL backbones.}
    \label{fig:comp}
    \vspace{-3mm}
\end{figure*}

Our method supports two distinct tasks: pixelization of existing images using only the second inference stage, and text-to-image pixel art generation using the complete two-stage pipeline. We evaluate both settings separately along two dimensions:
(i) perceptual image quality, and 
(ii) computational efficiency.

\subsection{PAPify Setup}

All experiments use Stable Diffusion 3.5 Medium as the backbone model, except for comparisons against SD-$\pi$XL and VectorFusion, where we use SDXL for a fair comparison. The second inference is initialized at the unshifted timestep $t_s=0.6$, so its initial noise level $\sigma_s=\sigma(t_s;m)$ adapts to the chosen flow shift $m$.
Given a target number of colors, we use k-means~\cite{macqueen1967kmeans} on the reference image to obtain the palette used throughout the second inference stage.
All pixelization experiments are performed on an NVIDIA Tesla T4 GPU with 15 GB of available VRAM. For faster execution, text-to-image generation experiments are instead conducted on an NVIDIA A100 GPU.
All inference parameters of PAPify, along with the evaluation prompts and random seeds, are provided in the supplementary material.


\subsection{Metrics}

We evaluate the aesthetic quality of the resulting pixel art, focusing on visual coherence and adherence to pixel art artistic conventions~\cite{gerstner2012pixelated,wu2022sprites}.

We adopt a pairwise comparison protocol inspired by the Visual Aesthetic Benchmark (VAB)~\cite{feng2026vab}, which investigates the ability of multimodal language models to assess visual aesthetics. For each comparison, GPT-6.1 Sol~\cite{openai2026gpt61sol} serves as an automated judge and selects the aesthetically preferable image according to pixel art-specific artistic criteria. Each pair contains an output from PAPify and one from a competing baseline, produced under the same target constraints. Image positions are randomized to mitigate positional bias. We report PAPify's win rate against each baseline, defined as the percentage of comparisons in which its output is preferred.

We exclude CLIPScore~\cite{hessel2021clipscore} because preliminary experiments showed only marginal differences across methods, limiting its usefulness in our setting. Our quantitative evaluation therefore focuses on aesthetic preference rather than semantic alignment.


\subsection{Generation}

\paragraph{Baselines.}
For text-to-image pixel art generation, we compare our complete pipeline against three generative models: GPT-Image-2.5 Sunburst (\texttt{quality=low})~\cite{openai2026gptimage25sunburst}, FLUX.2 [klein] 4B~\cite{blackforestlabs2026flux2klein4b}, and Stable Diffusion 3.5 Medium~\cite{stabilityai2024sd35}. The latter serves as our base model, making this comparison particularly important for assessing whether our method improves generation quality beyond the capabilities of the underlying model.

\paragraph{Protocol.}
We evaluate generation using 100 prompts randomly sampled from GenAI-Bench~\cite{li2024genaibench}, reusing the same prompts across three pixel art resolutions: $48\times48$, $64\times64$, and $128\times128$, with at most 10, 16, and 32 colors, respectively.
Each prompt is augmented with a request to generate pixel art at the specified resolution and within the corresponding color limit. All methods receive identical inputs.
We apply PAP to the outputs of competing models to enforce the target spatial and color constraints.
We assess aesthetic quality using VAB-based pairwise comparisons between PAPify and each baseline. Results are reported in the Generation section of Table~\ref{tab:pairwise_winrate}.

\subsection{Pixelization}

\paragraph{Baselines.}
For image pixelization, we compare our method against seven baselines: Pixelated Image Abstraction (PIA)~\cite{gerstner2012pixelated}, GPT-Image-2.5 Sunburst~\cite{openai2026gptimage25sunburst}, direct Pixel Art Projection (PAP), FLUX.2 [klein] 4B~\cite{blackforestlabs2026flux2klein4b}, Art-Oriented Pixelation (AOP)~\cite{lei2024art}, Automatic Portrait Image Pixelization (APIP)~\cite{shang2021automatic}, and Make Your Own Sprites (MYOS)~\cite{wu2022sprites}.

\paragraph{Protocol.}
We generate reference images with Stable Diffusion 3.5 Medium using the same 100 randomly sampled GenAI-Bench prompts~\cite{li2024genaibench} as in the generation benchmark. Each reference image is pixelized by all seven baselines and PAPify at three pixel art resolutions: $48\times48$, $64\times64$, and $128\times128$, with at most 10, 16, and 32 colors, respectively.
PIA, PAP, AOP, APIP, and MYOS are applied directly to the same reference images. GPTImage and FLUX receive the reference image and an identical prompt requesting pixelization at the target block size resolution within the corresponding color limit. We apply PAP to their outputs to enforce the spatial and color constraints. As in SD-$\pi$XL, PAPify uses both the reference image and its original text prompt as conditioning inputs.
We assess aesthetic quality using VAB-based pairwise comparisons between PAPify and each baseline. Results are reported in the Pixelization section of Table~\ref{tab:pairwise_winrate}.

\subsection{Comparison with Gradient-Based Methods}

\paragraph{Baselines.}
We compare PAPify against SD-$\pi$XL~\cite{binninger2024sdpixl} and VectorFusion~\cite{jain2023vectorfusion}, two gradient-based methods that use pretrained diffusion models for constrained image generation.

\paragraph{Protocol.}
We use the same evaluation benchmark as SD-$\pi$XL: 150 prompts from PartiPrompts~\cite{yu2022parti}, equally distributed across pixel art resolutions of $32\times32$, $48\times48$, and $64\times64$. All images are limited to 8 colors, regardless of resolution.
For a fair comparison, we implement PAPify with the same pretrained backbone, Stable Diffusion XL (SDXL), allowing us to assess our sampling procedure independently of backbone differences.
We assess aesthetic quality using VAB-based pairwise comparisons between PAPify and each baseline.


\subsection{Results}

As shown in Table~\ref{tab:pairwise_winrate}, our method outperforms all seven pixelization baselines on average, with win rates ranging from 55.0\% against GPTImage to 91.7\% against AOP. For text-to-image generation, our method achieves average win rates from 65.7\% against GPTImage to 93.3\% against SD3.5 Medium.

Although GPTImage outperforms our method at $128\times128$ resolution for pixelization, our approach remains competitive despite relying on a smaller generative model.

Importantly, our method is not tied to SD3.5 Medium and can be extended to other compatible diffusion models. The substantial improvements over SD3.5, with win rates of 87.0\% for pixelization against direct PAP and 93.3\% for generation against pixel art prompting followed by PAP, suggest that our approach could yield further improvements when applied to stronger generative models. This model independence is further supported by our implementation on SDXL, which achieves strong results against SD-$\pi$XL and VectorFusion, as shown in Table~\ref{tab:sdpixl_comparison}.

Interestingly, during generation, we observed that general-purpose image generators, including GPTImage, FLUX, and SD3.5, struggle to follow explicit pixel art resolution constraints. Despite requesting different grid sizes, these models tend to produce pixel art at similar apparent resolutions, often yielding nearly identical images across the three target block size resolutions. We observed similar difficulties in respecting the requested number of colors.

Overall, these results demonstrate that integrating PAP constraints during diffusion produces more aesthetically convincing pixel art than conventional pixelization and generation methods, while maintaining competitive inference times.

\begin{table}[t]
    \centering
    \small
    \setlength{\tabcolsep}{5pt}
    \begin{tabular}{lccccc}
        \toprule
        Opponent
        & 48
        & 64
        & 128
        & Avg.
        & Time $\downarrow$ \\
        \midrule

        \multicolumn{6}{l}{\textbf{Pixelization}} \\
        \midrule

        \textit{PAP}
        & \textcolor{green!60!black}{72\%}
        & \textcolor{green!60!black}{93\%}
        & \textcolor{green!60!black}{96\%}
        & \textcolor{green!60!black}{87.0\%}
        & $0.2\,\mathrm{s}$ \\

        AOP~\cite{lei2024art}
        & \textcolor{green!60!black}{80\%}
        & \textcolor{green!60!black}{98\%}
        & \textcolor{green!60!black}{97\%}
        & \textcolor{green!60!black}{91.7\%}
        & $0.3\,\mathrm{s}$ \\

        APIP~\cite{shang2021automatic}
        & \textcolor{green!60!black}{83\%}
        & \textcolor{green!60!black}{88\%}
        & \textcolor{green!60!black}{69\%}
        & \textcolor{green!60!black}{80.0\%}
        & $1\,\mathrm{s}$ \\

        MYOS~\cite{wu2022sprites}
        & \textcolor{green!60!black}{71\%}
        & \textcolor{green!60!black}{91\%}
        & \textcolor{green!60!black}{93\%}
        & \textcolor{green!60!black}{85.0\%}
        & $0.2\,\mathrm{s}$ \\

        PIA~\cite{gerstner2012pixelated}
        & \textcolor{green!60!black}{71\%}
        & \textcolor{green!60!black}{78\%}
        & \textcolor{green!60!black}{85\%}
        & \textcolor{green!60!black}{78.0\%}
        & $6\,\mathrm{s}$ \\

        FLUX~\cite{blackforestlabs2026flux2klein4b}
        & \textcolor{green!60!black}{91\%}
        & \textcolor{green!60!black}{92\%}
        & \textcolor{green!60!black}{80\%}
        & \textcolor{green!60!black}{87.7\%}
        & $6\,\mathrm{s}^{\dagger}$ \\

        GPTImage~\cite{openai2026gptimage25sunburst}
        & \textcolor{green!60!black}{62\%}
        & \textcolor{green!60!black}{58\%}
        & \textcolor{red!70!black}{45\%}
        & \textcolor{green!60!black}{55.0\%}
        & $21\,\mathrm{s}^{\dagger}$ \\

        \midrule
        \multicolumn{6}{l}{\textbf{Generation}} \\
        \midrule

        \textit{SD 3.5}~\cite{stabilityai2024sd35}
        & \textcolor{green!60!black}{88\%}
        & \textcolor{green!60!black}{93\%}
        & \textcolor{green!60!black}{99\%}
        & \textcolor{green!60!black}{93.3\%}
        & $30\,\mathrm{s}$ \\

        FLUX~\cite{blackforestlabs2026flux2klein4b}
        & \textcolor{green!60!black}{87\%}
        & \textcolor{green!60!black}{67\%}
        & \textcolor{gray}{51\%}
        & \textcolor{green!60!black}{68.3\%}
        & $6\,\mathrm{s}^{\dagger}$ \\

        GPTImage~\cite{openai2026gptimage25sunburst}
        & \textcolor{green!60!black}{76\%}
        & \textcolor{green!60!black}{71\%}
        & \textcolor{gray}{50\%}
        & \textcolor{green!60!black}{65.7\%}
        & $21\,\mathrm{s}^{\dagger}$ \\

        \bottomrule
    \end{tabular}
    \caption{Pairwise preference rates (\%) of PAPify against pixelization and generation baselines at three target block size resolutions. Avg. denotes the mean across resolutions. Inference times refer to the competing methods. $\dagger$ denotes API time.}
    
    \label{tab:pairwise_winrate}
\end{table}

\begin{table}[t]
    \centering
    \small
    \setlength{\tabcolsep}{5pt}
    \begin{tabular}{lccccc}
        \toprule
        Opponent
        & 32
        & 48
        & 64
        & Avg.
        & Time $\downarrow$ \\
        \midrule

        SD-$\pi$XL~\cite{binninger2024sdpixl}
        & \textcolor{green!60!black}{60\%}
        & \textcolor{green!60!black}{66\%}
        & \textcolor{green!60!black}{60\%}
        & \textcolor{green!60!black}{62\%}
        & $1\,\mathrm{h}\,30\,\mathrm{min}$
        \\

        VectorFusion~\cite{jain2023vectorfusion}
        & \textcolor{green!60!black}{90\%}
        & \textcolor{green!60!black}{80\%}
        & \textcolor{green!60!black}{100\%}
        & \textcolor{green!60!black}{90\%}
        & $25\,\mathrm{min}$
        \\

        \bottomrule
    \end{tabular}
    \caption{Pairwise preference rates (\%) of PAPify against SD-$\pi$XL and VectorFusion on the SD-$\pi$XL benchmark. All PAPify results use SDXL as the backbone for a controlled comparison. Inference times refer to the competing methods.}
    \label{tab:sdpixl_comparison}
\end{table}

\subsection{Fidelity--Aesthetics Trade-off}
\label{sec:constraint_quality}

Pixel art generation involves a trade-off between preserving fidelity to the reference image and adapting its content to pixel art artistic conventions. PAPify allows users to adjust this balance through its sampling parameters and the target spatial and color constraints.

\begin{figure}[!t]
    \centering
    \makebox[\columnwidth][c]{%
    \includegraphics[
        width=1\columnwidth,
        keepaspectratio
    ]{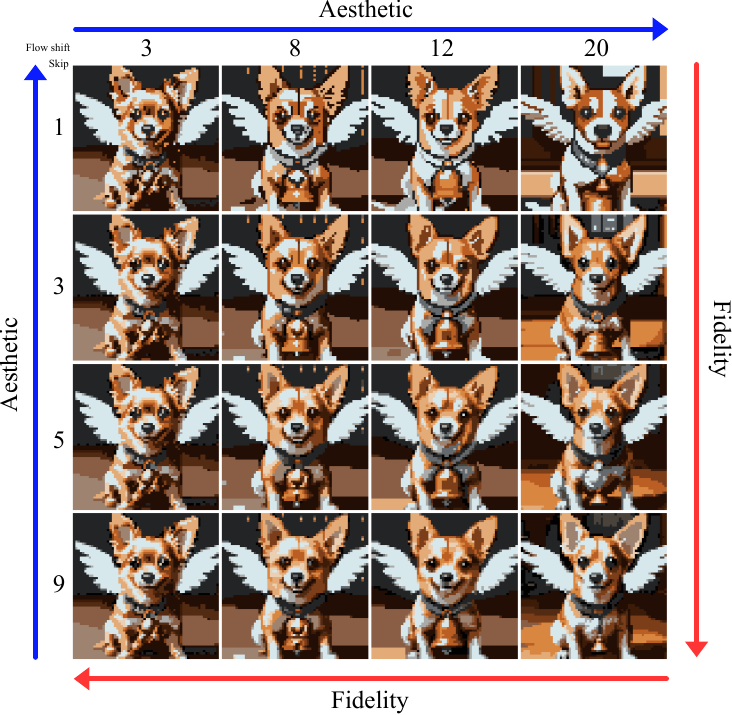}%
    }
    \caption{Effect of $k_{\mathrm{skip}}$ and flow shift on pixel art synthesis. Their combination controls the degree of stylistic transformation and preservation of the reference image.}
    \label{fig:skip_flowshift_figure}
\end{figure}

\paragraph{$\boldsymbol{k_{\mathrm{skip}}}$.}
$k_{\mathrm{skip}}$ controls how often projection is applied during denoising. Larger values favor fidelity to the reference composition by applying fewer PAP-constrained updates. Smaller values encourage stronger stylization, with simplified shapes, sharper contours, and more coherent pixel clusters, potentially introducing greater deviations from the reference image.
As illustrated in Fig.~\ref{fig:skip_flowshift_figure}, frequent PAP updates combined with a higher flow shift encourage more substantial changes to the scene's structure and composition.

\begin{figure}[t]
    \centering
    \makebox[\columnwidth][c]{%
        \includegraphics[
            width=1\columnwidth,
            keepaspectratio
        ]{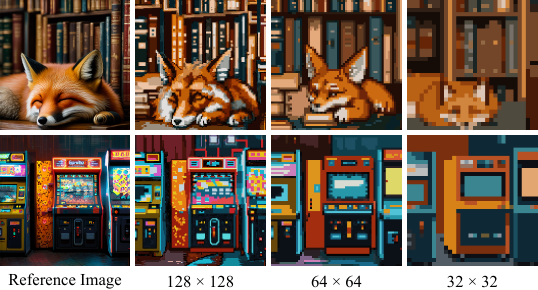}
    }
    \caption{Resolution-dependent adaptation of PAPify. As the target block size resolution decreases, the model progressively simplifies object shapes and scene composition to maintain visual readability.}
    \label{fig:resolution_adaptation}
\end{figure}

\paragraph{Target Block Size Resolution.}
The target block size resolution $p$ determines how much visual information can be represented. As shown in Fig.~\ref{fig:resolution_adaptation}, PAPify adapts both the level of detail and the semantic representation of the scene to this constraint. At lower pixel art resolutions, it simplifies object shapes and reorganizes scene elements to maintain a coherent and recognizable composition rather than attempting to retain details that cannot be meaningfully represented.

\paragraph{Color Palette.}
Fig.~\ref{fig:palette} illustrates a similar adaptation to the target color palette $c$. A limited palette provides fewer color distinctions for separating objects and conveying details. PAPify adapts the visual structure accordingly, preserving a coherent and recognizable composition within the available colors.

\section{Ablation Study}
\label{sec:ablations}

We investigate the contribution of three key components of our method: the initial reference-image generation stage, the Full PAP Update, and the flow shift adaptation. Each component is independently removed while keeping the remaining configuration unchanged. We evaluate the resulting variants using the same pairwise preference protocol described in Sec.~\ref{sec:results}.
\begin{table}[t]
    \centering
    \small
    \setlength{\tabcolsep}{6pt}
    \begin{tabular}{lcccc}
        \toprule
        Ablation
        & 48 & 64 & 128 & Avg. \\
        \midrule

        w/o First Inference
        & \textcolor{green!60!black}{59\%}
        & \textcolor{green!60!black}{57\%}
        & \textcolor{red!75!black}{44\%}
        & \textcolor{green!60!black}{53.3\%} \\

        w/o Full PAP Update
        & \textcolor{green!60!black}{84\%}
        & \textcolor{green!60!black}{88\%}
        & \textcolor{green!60!black}{90\%}
        & \textcolor{green!60!black}{87.3\%} \\

        w/o Flow Shift
        & \textcolor{green!60!black}{75\%}
        & \textcolor{green!60!black}{89\%}
        & \textcolor{green!60!black}{86\%}
        & \textcolor{green!60!black}{83.3\%} \\

        \bottomrule
    \end{tabular}
    \caption{
    Ablation study reporting the pairwise win rate
    of our full configuration against each ablated
    variant across target block size resolutions.
    }
    \label{tab:ablation_winrate}
\end{table}
As shown in Table~\ref{tab:ablation_winrate}, removing each key component independently degrades the aesthetic quality of the pixel art. These findings highlight the importance of diffusion-time PAP constraints and flow shift adaptation, while suggesting that the benefits of the initial generation stage depend on the target block size resolution.

%% file: sec/6_CONCLUSION.tex
\section{Conclusion}

In this work, we introduced PAPify, a training-free method for generating pixel art with explicit spatial and color constraints using pretrained diffusion models. By integrating Pixel Art Projection directly into the denoising process, our approach allows the model to adapt its composition and visual structure to the target pixel art resolution, rather than relying on post-processing alone. PAPify can also be adapted to other diffusion models that use the same sampling framework or compatible variants.

Our experiments demonstrate that PAPify consistently outperforms existing pixelization and generation baselines in aesthetic quality, while remaining computationally efficient. We further show that adapting the flow shift to the target pixel size improves denoising behavior, and that projection frequency provides control over the trade-off between reference fidelity and pixel art aesthetics.

More broadly, our framework is not limited to the specific PAP operator used in this work for pixel art. Future research could explore more sophisticated mathematical PAP operators to further improve pixel art quality, as well as other transformations, such as polygonization or artistic filtering, to guide pretrained diffusion models toward different artistic domains. This suggests a broader direction for training-free artistic specialization through the integration of mathematical transformations directly into the diffusion process.

%% file: sec/7_ACKNOWLEDGMENTS.tex
\section{Acknowledgement}

We thank Stéphanie de Senneville for creating the embroidery shown in Figure 2 and Armand de Senneville for his support throughout the project.

%% file: sec/8_LLM_USAGE.tex
\section*{LLM Usage Statement}

We used OpenAI Codex and ChatGPT to assist with code development, including generating small code snippets, plotting, reviewing author-written code, implementing methods, adapting code from existing repositories, and conducting exploratory experiments. These tools also supported drafting and revising the manuscript, language editing, and discussions of results and their interpretation. They also assisted with literature searches for the Related Work section.

GPT-6.1 Sol served as an automated judge for the pairwise aesthetic comparisons reported in the experiments and ablation study.

The authors retain full responsibility for the implementation, experimental protocol, reported results, and scientific claims.

%% file: sec/X_suppl.tex
\onecolumn
\appendix
\setcounter{page}{1}
\renewcommand{\thepage}{S\arabic{page}}
\providecommand{\mainpapersection}[2]{Sec.~\ref{#1}}
\raggedbottom
\ifdefined\nolinenumbers
    \nolinenumbers
\fi
\pdfbookmark[0]{Supplementary Material}{supplementary-material}
\begin{center}
    \Large
    \textbf{\thetitle}\par
    \vspace{0.5em}
    Supplementary Material
\end{center}

\counterwithin{figure}{section}
\renewcommand{\thefigure}{\thesection\arabic{figure}}
\renewcommand{\theHfigure}{supp.\thesection.\arabic{figure}}
\counterwithin{table}{section}
\renewcommand{\thetable}{\thesection\arabic{table}}
\renewcommand{\theHtable}{supp.\thesection.\arabic{table}}

\input{sec/supp/A_implementation}

\clearpage
\section{Visual Results}
\label{sec:supp_visual_results}

We present qualitative results for PAPify across a range of prompts, output resolutions, and palette settings.

\begin{center}
    \begin{minipage}{\textwidth}
        \centering
        \includegraphics[width=\linewidth]{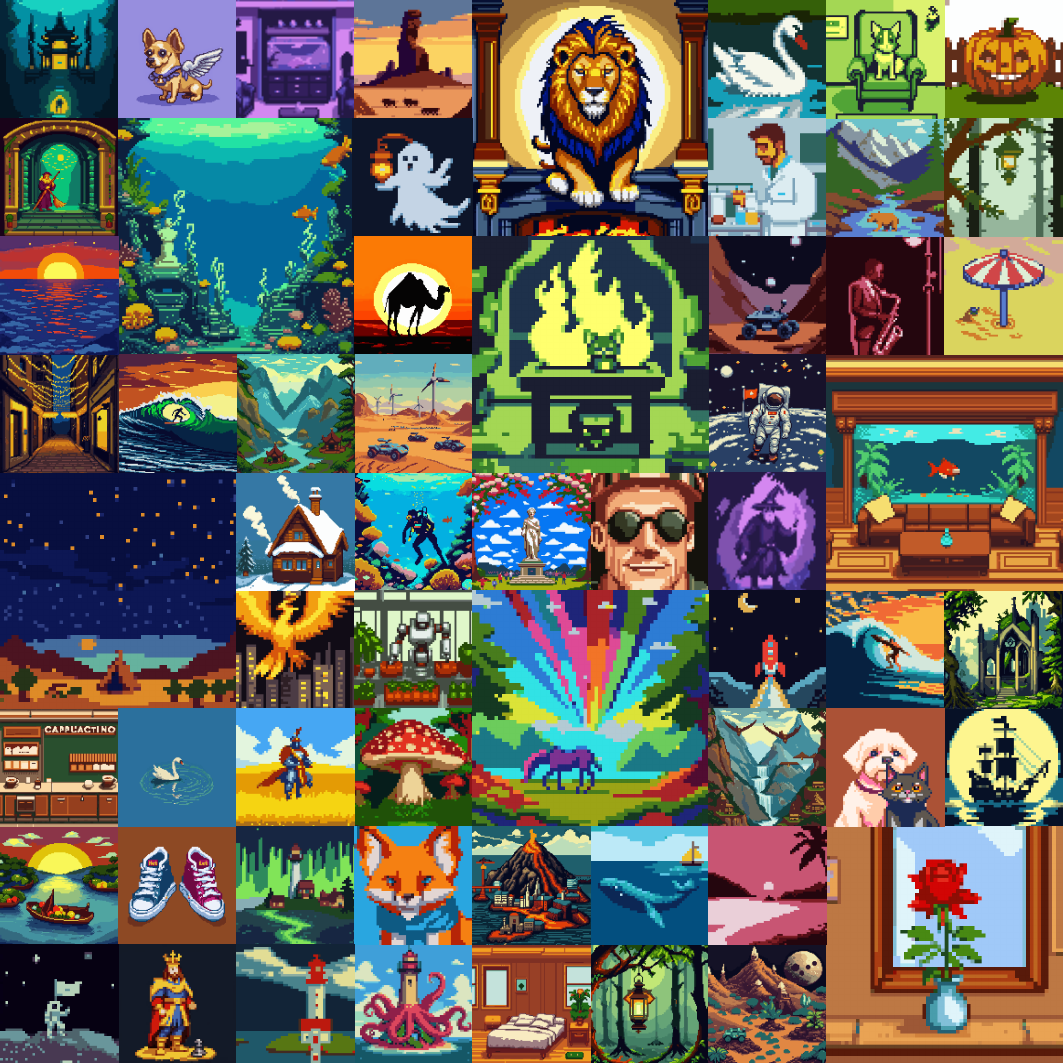}
        \captionsetup{hypcap=false}
        \captionof{figure}{\textbf{Qualitative results.}
        Selected PAPify outputs across prompts, target grid sizes, and palette settings.}
        \label{fig:supp_qualitative_gallery}
    \end{minipage}
\end{center}
\clearpage

\newsavebox{\suppPaletteCaption}
\begin{figure}[p]
    \centering
    \captionsetup{position=bottom}
    \setbox\suppPaletteCaption=\vbox{%
        \caption{\textbf{Palette control at a fixed resolution.}
        Each row uses the same text prompt and a $64\times64$ grid.}
        \label{fig:supp_palette_gallery}%
    }
    \includegraphics[
        width=\textwidth,
        height=\dimexpr\textheight-\ht\suppPaletteCaption-\dp\suppPaletteCaption-1pt\relax,
        keepaspectratio
    ]{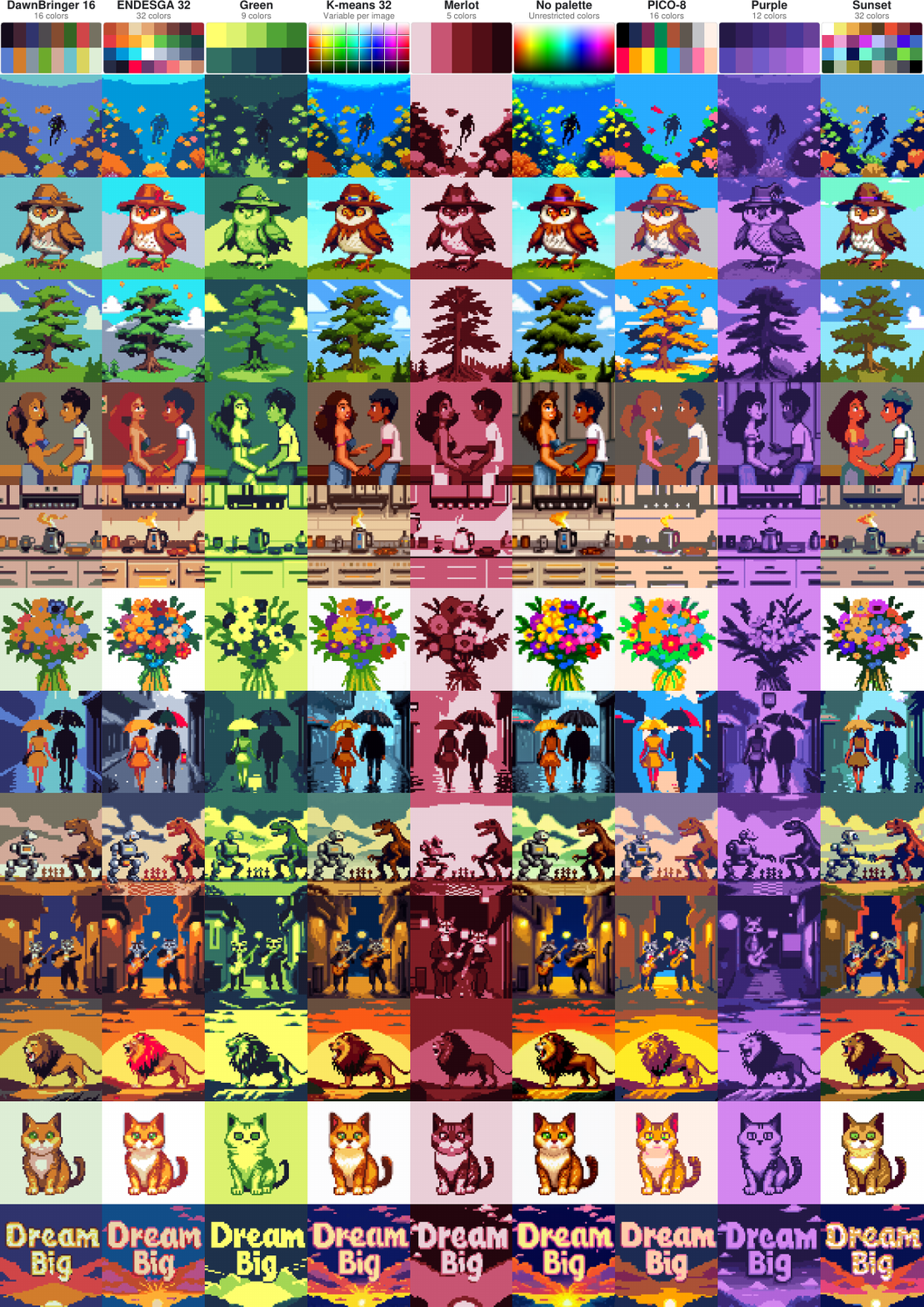}\par
    \nointerlineskip
    \box\suppPaletteCaption
\end{figure}
\clearpage

\input{sec/supp/B_palettes}
\input{sec/supp/C_evaluation}

\input{sec/supp/D_compute}
\input{sec/supp/E_analyses}
\input{sec/supp/F_results_limits}

%% file: sec/supp/A_implementation.tex
\section{Reproducible Implementation}
\label{sec:supp_implementation}

We use SD3.5 Medium for the main generation and pixelization experiments and SDXL for the comparison with gradient-based generation methods.

\subsection{Stable Diffusion 3.5 Medium}
\label{sec:supp_parameters}

Table~\ref{tab:supp_sd35_config} lists the SD3.5 Medium configuration used by PAPify.

\par\smallskip\noindent
\begin{minipage}{\columnwidth}
    \centering
    \small
    \begin{tabular}{@{}p{0.26\linewidth}p{0.70\linewidth}@{}}
        \toprule
        Setting & Value \\
        \midrule
        Backbone and VAE & SD3.5 Medium~\cite{stabilityai2024sd35} and its VAE in FP16 \\
        Text encoders & CLIP-L and CLIP-G in FP32 on CPU \\
        T5-XXL & Disabled with zero conditioning \\
        Negative prompt & Empty \\
        Base sampler & Euler, 50 steps, classifier-free guidance (CFG) scale 5 (flow shift~3 for the first-stage inference) \\
        Skip Layer Guidance & Layers 7--9, scale 2.5, active at 1--20\% of the first-stage schedule \\
        \bottomrule
    \end{tabular}
    \captionsetup{hypcap=false,skip=5pt}
    \captionof{table}{Stable Diffusion 3.5 Medium configuration for the baseline and the PAPify backbone.}
    \label{tab:supp_sd35_config}
\end{minipage}

\subsection{PAPify Generation and Pixelization}

The RGB canvas is the image on which diffusion operates. A square canvas of side length $L$ and a PAP block width $p$ define an $R\times R$ output grid with $R=L/p$. The palette settings are described in Sec.~\ref{sec:supp_palettes}.

\paragraph{Generation.}
Both stages use the same prompt (Sec.~\ref{sec:supp_prompts}). With the chosen configuration, the second inference stage always runs for 17 steps. Table~\ref{tab:supp_papify_generation} lists the PAPify settings used to generate the pixel art in our results.

\par\smallskip\noindent
\begin{minipage}{\columnwidth}
    \centering
    \small
    \begin{tabular*}{\linewidth}{@{\extracolsep{\fill}}lcccc@{}}
        \toprule
        Target grid & RGB canvas & Block width $p$ & Maximum colors $K$ & Flow shift \\
        \midrule
        $48\times48$  & $384\times384$  & 8 & 10 & 12 \\
        $64\times64$  & $512\times512$  & 8 & 16 & 12 \\
        $128\times128$ & $1024\times1024$ & 8 & 32 & 12 \\
        \midrule
        \multicolumn{5}{@{}l@{}}{\textit{Shared settings}} \\
        \multicolumn{2}{@{}l}{First inference flow shift}
        & \multicolumn{3}{l@{}}{3} \\
        \multicolumn{2}{@{}l}{Initial noise level for refinement}
        & \multicolumn{3}{l@{}}{$t=0.6$} \\
        \multicolumn{2}{@{}l}{Sampling}
        & \multicolumn{3}{l@{}}{50 nominal steps, CFG scale 5} \\
        \multicolumn{2}{@{}l}{$k_{\mathrm{skip}}$}
        & \multicolumn{3}{l@{}}{5} \\
        \multicolumn{2}{@{}l}{Palette extraction}
        & \multicolumn{3}{l@{}}{FP32 k-means, 20 iterations, diversity 0.8} \\
        \bottomrule
    \end{tabular*}
    \captionsetup{hypcap=false,skip=5pt}
    \captionof{table}{PAPify text-to-image generation with SD3.5 Medium. The fixed seed is 2.}
    \label{tab:supp_papify_generation}
\end{minipage}

\paragraph{Pixelization.}
Pixelization starts from a saved reference image and uses its original text description to condition refinement. Table~\ref{tab:supp_papify_pixelization} lists the grid-specific and shared refinement settings used to obtain the reported pixelization results.

\par\smallskip\noindent
\begin{minipage}{\columnwidth}
    \centering
    \small
    \begin{tabular*}{\linewidth}{@{\extracolsep{\fill}}lcccc@{}}
        \toprule
        Target grid & RGB canvas & Block width $p$ & Maximum colors $K$ & Flow shift \\
        \midrule
        $48\times48$  & $384\times384$ & 8 & 10 & 12 \\
        $64\times64$  & $512\times512$ & 8 & 16 & 12 \\
        $128\times128$ & $512\times512$ & 4 & 32 & 6  \\
        \midrule
        \multicolumn{5}{@{}l@{}}{\textit{Shared settings}} \\
        \multicolumn{2}{@{}l}{Reference flow shift for $t$}
        & \multicolumn{3}{l@{}}{3} \\
        \multicolumn{2}{@{}l}{Initial noise level for refinement}
        & \multicolumn{3}{l@{}}{$t=0.6$} \\
        \multicolumn{2}{@{}l}{Sampling}
        & \multicolumn{3}{l@{}}{50 nominal steps, CFG scale 5} \\
        \multicolumn{2}{@{}l}{$k_{\mathrm{skip}}$}
        & \multicolumn{3}{l@{}}{5} \\
        \multicolumn{2}{@{}l}{Palette extraction}
        & \multicolumn{3}{l@{}}{FP32 k-means, 20 iterations, diversity 0.8} \\
        \bottomrule
    \end{tabular*}
    \captionsetup{hypcap=false,skip=5pt}
    \captionof{table}{PAPify pixelization settings with SD3.5 Medium. The fixed seed is 23.}
    \label{tab:supp_papify_pixelization}
\end{minipage}

\clearpage
\subsection{Stable Diffusion XL}
\label{sec:supp_sdxl}

For the preference comparison with SD-$\pi$XL and VectorFusion, we use the SDXL configuration in Table~\ref{tab:supp_sdxl_config}. Table~\ref{tab:supp_papify_sdxl} gives the PAPify generation settings for the three benchmark grids. The runtime profiling configuration is given in Sec.~\ref{sec:supp_compute}.

\par\smallskip\noindent
\begin{minipage}{\columnwidth}
    \centering
    \small
    \begin{tabular}{@{}p{0.26\linewidth}p{0.70\linewidth}@{}}
        \toprule
        Setting & Value \\
        \midrule
        Backbone & SDXL Base 1.0~\cite{podell2024sdxl} in FP16 \\
        VAE & TAESDXL in FP16 \\
        Sampler & Euler discrete \\
        Sampling steps & 40 \\
        CFG scale & 7.5 \\
        Negative prompt & Empty \\
        Seed & 42 (fixed) \\
        \bottomrule
    \end{tabular}
    \captionsetup{hypcap=false,skip=5pt}
    \captionof{table}{SDXL configuration used by PAPify.}
    \label{tab:supp_sdxl_config}
\end{minipage}

\par\smallskip\noindent
\begin{minipage}{\columnwidth}
    \centering
    \small
    \begin{tabular*}{\linewidth}{@{\extracolsep{\fill}}lccc@{}}
        \toprule
        Target grid & RGB canvas & Block width $p$ & Maximum colors $K$ \\
        \midrule
        $32\times32$ & $768\times768$ & 24 & 8 \\
        $48\times48$ & $768\times768$ & 16 & 8 \\
        $64\times64$ & $768\times768$ & 12 & 8 \\
        \midrule
        \multicolumn{4}{@{}l@{}}{\textit{Shared settings}} \\
        \multicolumn{2}{@{}l}{First inference start}
        & \multicolumn{2}{l@{}}{$t=1.0$} \\
        \multicolumn{2}{@{}l}{Refinement interval}
        & \multicolumn{2}{l@{}}{$t=0.7$ to $t=0$} \\
        \multicolumn{2}{@{}l}{$k_{\mathrm{skip}}$}
        & \multicolumn{2}{l@{}}{5} \\
        \multicolumn{2}{@{}l}{PAP update boost $k_{\mathrm{boost}}$}
        & \multicolumn{2}{l@{}}{10} \\
        \multicolumn{2}{@{}l}{Palette extraction}
        & \multicolumn{2}{l@{}}{20 k-means iterations, diversity 0.8} \\
        \bottomrule
    \end{tabular*}
    \captionsetup{hypcap=false,skip=5pt}
    \captionof{table}{PAPify text-to-image generation with SDXL. Each image uses its own palette throughout refinement and final projection.}
    \label{tab:supp_papify_sdxl}
\end{minipage}

%% file: sec/supp/B_palettes.tex
\section{Palettes and Control}
\label{sec:supp_palettes}

We describe the three color settings that PAPify supports. Users can leave colors unconstrained, supply a palette, or specify a maximum number of colors.

\paragraph{No palette.}
With palette quantization disabled, PAP replaces the pixels in each $p\times p$ block with their mean RGB color. Even without an explicit color constraint, PAPify tends toward common pixel-art conventions~\cite{gerstner2012pixelated} during refinement. A few dominant colors define shapes and lighting, and sharp outlines separate objects from their backgrounds. Figure~\ref{fig:supp_palette_modes} illustrates both traits across all three grids.

\begin{center}
    \includegraphics[width=0.87\textwidth]{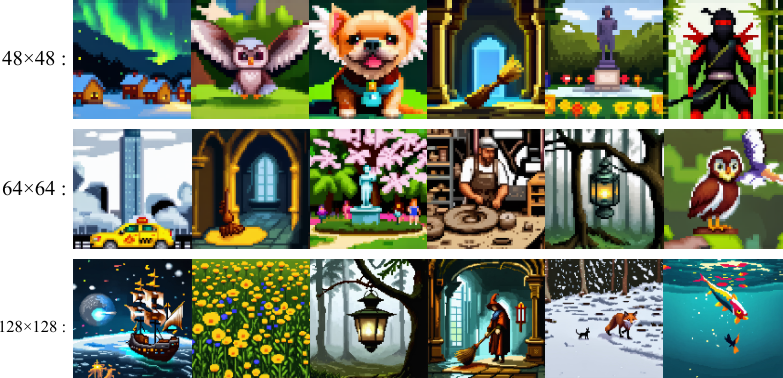}
    \captionsetup{hypcap=false}
    \captionof{figure}{\textbf{PAPify without an imposed palette.} Examples at three output resolutions.}
    \label{fig:supp_palette_modes}
\end{center}

\paragraph{Fixed palette.}
For each block mean $a$, PAP selects the palette color $b$ that minimizes our color-matching score

\begin{equation}
    d(a,b) = \frac{\lVert a-b\rVert_2}{255} + |S(a)-S(b)| + 4 S(a) S(b) \min(\Delta H,1-\Delta H),
    \label{eq:supp_color_distance}
\end{equation}

where RGB values lie in $[0,255]$, $H$ and $S$ are HSV hue and saturation~\cite{smith1978color} in $[0,1]$, and $\Delta H=|H(a)-H(b)|$. The hue term uses the shorter arc on the hue circle. Hue differences are less apparent at low saturation~\cite{sural2002hsv}, so we weight the hue term by the product of the two saturations.

\paragraph{Maximum number of colors.}
\label{sec:supp_palette}

We extract a palette of up to $K$ colors from the reference image's block means using weighted RGB $k$-means~\cite{macqueen1967kmeans}. We initialize the centers from evenly spaced samples in luminance order and run 20 Lloyd iterations~\cite{lloyd1982quantization}. We round the centers to 8-bit RGB and keep the palette fixed throughout refinement and export.

PAPify's \emph{diversity} parameter $d$ gives more influence to rare colors, which frequency-based quantization can overlook~\cite{heckbert1982color}. We group block colors into RGB bins of width 32 per channel and give each sample in a bin of $n$ samples weight $n^{-d}$. At $d=0$, all samples have the same weight. At $d=1$, all occupied bins have the same total weight. We use $d=0.8$.

%% file: sec/supp/C_evaluation.tex
\section{Evaluation Protocol}
\label{sec:supp_evaluation}

\subsection{Aesthetic Evaluation}

Following the comparative design of the Visual Aesthetic Benchmark (VAB)~\cite{feng2026vab}, we ask a multimodal judge to choose the aesthetically stronger pixel-art image in each pair. Both images in a pair share the same target pixel-art grid and are presented at the same display size. Section~\ref{sec:supp_judge_prompt} reproduces the instructions sent to the judge.

\paragraph{Pairing and scoring.}
GPT-6.1 Sol (\texttt{openai/gpt-6.1-sol})~\cite{openai2026gpt61sol} receives the benchmark PNGs through OpenRouter with image detail \texttt{auto}, medium reasoning effort, and a completion limit of 8192 tokens. We randomize the A/B order with seed 23 and request one judgment per prompt, opponent, and resolution. PAPify's win rate is the fraction of pairs in which it is preferred. The aggregate win rate is the arithmetic mean of the three resolution-specific rates.

\begin{samepage}
    \input{sec/supp/C_judge_prompt}
\end{samepage}

\input{sec/X_supp_judge_rationale}

\subsection{Prompts and Baseline Settings}
\label{sec:supp_prompts}

\paragraph{Prompt selection.}
We use the same 100 descriptions randomly sampled from GenAI-Bench~\cite{li2024genaibench} for both generation and pixelization. We reuse them at $48\times48$, $64\times64$, and $128\times128$, with color budgets of 10, 16, and 32, respectively. The SDXL comparison uses the 150 PartiPrompts descriptions~\cite{yu2022parti} from the SD-$\pi$XL benchmark~\cite{binninger2024sdpixl}, with 50 descriptions at each of $32\times32$, $48\times48$, and $64\times64$ and a budget of eight colors per image. We compare PAPify against the $k$-means palette variant of SD-$\pi$XL and against VectorFusion.

\paragraph{Generation prompt.}
PAPify, prompted SD3.5 Medium, GPTImage, and FLUX receive the same text template. The fields \texttt{prompt}, \texttt{size}, and \texttt{n\_colors} denote the original scene description, the side length of the square output grid, and the maximum number of colors, respectively.

\begin{quote}
    \small\raggedright\ttfamily
    \{prompt\}, pixel art on a \{size\}x\{size\} pixel grid, at most \{n\_colors\} colors
\end{quote}

\paragraph{Pixelization prompt.}
GPTImage and FLUX receive the reference image and the instruction below. The fields \texttt{size} and \texttt{n\_colors} specify the target grid side length and the evaluation color budget.

\begin{quote}
    \small\raggedright\ttfamily
    Convert this image into pixel art on a \{size\}x\{size\} grid with \{n\_colors\} colors.
\end{quote}

PAPify receives the reference image and its original prompt, with the grid and color budget supplied directly to the projection operator.

\paragraph{API models.}
We request one PNG per call through OpenRouter, with provider fallbacks disabled. For GPT-Image-2.5 Sunburst~\cite{openai2026gptimage25sunburst}
(\texttt{openai/gpt-image-2.5-sunburst}), we request $1024\times1024$ images with the low-quality setting and an opaque background, using OpenAI as the sole provider. For FLUX.2 [klein] 4B~\cite{blackforestlabs2026flux2klein4b}
(\texttt{black-forest-labs/\allowbreak flux.2-klein-4b}), we request $512\times512$ images. For pairwise evaluation, we match the target pixel-art grid and color budget used for PAPify. We resize the GPTImage and FLUX results to the dimensions of PAPify's RGB inference canvas for the same task and target grid (Tables~\ref{tab:supp_papify_generation} and~\ref{tab:supp_papify_pixelization}).

%% file: sec/supp/C_judge_prompt.tex
\subsubsection{Aesthetic Judge Instructions}
\label{sec:supp_judge_prompt}

\begin{quote}
\small
You are judging PIXEL ART. Compare the two images labeled A and B and select the image with the best aesthetic quality as pixel art.

Prioritize pixel-level craftsmanship: readable silhouettes and essential features, coherent pixel clusters and color regions, clean contours with controlled stair-step patterns, and purposeful color choices. Favor selective abstraction that communicates forms clearly at limited resolution.

Penalize distracting pixel noise, fragmented shading, broken contours, and blurred transitions that weaken readability. Photographic detail, painterly richness, or smooth rendering must not outweigh coherent pixel design. Pixelation or color quantization alone does not establish quality.

Evaluate composition, color harmony, and artistic expression within this pixel-art standard. Simplicity and few colors are valid, but do not automatically indicate quality. Dithering and selective pixel-level anti-aliasing are acceptable when they support form or texture while preserving clarity.

Compare the visible results directly. Ignore labels, presentation order, presumed generation method, text-image alignment, and exact palette/resolution compliance.

You may explain your judgment before giving your decision. Your final line must be exactly \texttt{"BEST: A"} if image A is better, or \texttt{"BEST: B"} if image B is better. Choose exactly one winner, without ties. Any explanation must appear before this final line.
\end{quote}

%% file: sec/X_supp_judge_rationale.tex
\subsubsection{Design Rationale for the Aesthetic Judge Prompt}
\label{sec:supp_judge_rationale}

\paragraph{Selective abstraction and shape recognition.}
In \emph{Pixelated Image Abstraction}, Gerstner et al.~\cite{gerstner2012pixelated} jointly optimize which features survive downsampling and which palette colors represent them. Kopf et al.~\cite{kopf2013content} adapt downsampling kernels to local image structure, and Lei et al.~\cite{lei2024art} design pixelation around the salient features of cartoons. Biederman~\cite{biederman1987recognition} identifies contour properties as cues to object structure, and Gooch et al.~\cite{gooch2004human} show that facial illustrations can preserve recognition by emphasizing characteristic features. These studies motivate judging silhouettes and details by how well they communicate form at the target grid.

\paragraph{Contour continuity and stair steps.}
Kopf and Lischinski~\cite{kopf2011depixelizing} show why pixel connectivity matters for thin sprite features. Inglis and Kaplan~\cite{inglis2012pixelating} develop rasterization rules that reduce line artifacts, Kuo et al.~\cite{kuo2016feature} refine feature lines and their junctions in pixel-art animation, and Wu et al.~\cite{wu2022sprites} address aliasing during pixelization. Perceptual experiments by Field et al.~\cite{field1993contour} and Elder and Zucker~\cite{elder1993closure} show how alignment and closure help viewers read boundaries. The prompt therefore rewards step patterns that preserve the contour and penalizes breaks or blur that make it harder to follow.

\paragraph{Pixel clusters, shading, and noise.}
Palmer and Rock~\cite{palmer1994connectedness} propose uniform connectedness as a principle of perceptual grouping. Their account explains why contiguous pixels of similar color can read as a surface or shadow. Rosenholtz et al.~\cite{rosenholtz2007clutter} link local variation in color, contrast, and orientation to perceived clutter and difficulty in visual search. The prompt accordingly favors coherent clusters and penalizes noise or fragmented shading when either interrupts form or competes with the subject.

\paragraph{Color, composition, and expression.}
Ou and Luo~\cite{ou2006colour} model perceived harmony from ratings of color pairs. Schloss and Palmer~\cite{schloss2011aesthetic} distinguish harmony from preference for a particular figure against its background. For composition, Palmer et al.~\cite{palmer2008spatial} study the placement and facing direction of single objects, and Leyssen et al.~\cite{leyssen2012multiobject} study spacing and orientation in scenes with multiple objects. These results motivate assessing how colors separate regions and convey lighting, and how the arrangement and treatment of forms contribute to the whole image. The same criteria apply across the color budgets we evaluate.

\paragraph{Dithering and antialiasing.}
Heckbert~\cite{heckbert1982color} shows how dithering can reduce visible contouring with a limited palette, and Ulichney~\cite{ulichney1993void} designs dither arrays that avoid conspicuous patterns. Inglis et al.~\cite{inglis2013rasterizing} incorporate selective pixel-art antialiasing into \emph{Superpixelator} while preserving shape features such as sharp angles. The judge considers whether these techniques improve texture and tonal transitions while keeping contours readable.

\paragraph{Comparative aesthetic assessment.}
\emph{Pick-a-Pic} collects preferences between images generated for the same prompt~\cite{kirstain2023pick}. VAB studies expert comparisons of images with matched subject matter and controlled variation in rendering~\cite{feng2026vab}. The \mbox{SD-$\pi$XL} perceptual study reports separate rankings for aesthetics, prompt similarity, and reference fidelity~\cite{binninger2024sdpixl}, and HEIM treats aesthetics and text alignment as distinct evaluation dimensions~\cite{lee2023heim}. We present anonymous outputs at a shared target grid and color budget and ask the judge to choose based on their quality as pixel art. Studies of automated judging by Chen et al.~\cite{chen2024mllmjudge} and Wang et al.~\cite{wang2024fair} motivate randomizing the A/B order. The final \texttt{BEST} line supplies a single choice for the win-rate calculation.

%% file: sec/supp/D_compute.tex
\captionsetup{hypcap=false}

\section{Runtime and Memory}
\label{sec:supp_compute}

\subsection{PAPify with SD3.5 Medium}

For each configuration, we run PAPify with SD3.5 Medium on ten inputs, one at a time. Table~\ref{tab:supp_costs} reports the average runtime and its standard deviation, along with the peak GPU memory PyTorch uses (Alloc.) and reserves (Reserv.). The GPUs are a Tesla T4 and an A100-SXM4 with 80\,GB of memory.

\begin{center}
    \centering
    \small
    \setlength{\tabcolsep}{3pt}
    \begin{tabular}{@{}lccrrr@{}}
        \toprule
        GPU & Grid & RGB canvas & Time (s) & Alloc. & Reserv. \\
        \midrule
        \multicolumn{6}{@{}l}{\textit{Text-to-image generation}} \\
        T4   & $48\times48$  & $384\times384$  & $20.79\pm0.79$ & 5.25 & 5.51 \\
        T4   & $64\times64$  & $512\times512$  & $32.51\pm0.31$ & 5.60 & 6.01 \\
        T4   & $128\times128$ & $1024\times1024$ & $150.55\pm0.64$ & 7.97 & 9.30 \\
        A100 & $128\times128$ & $1024\times1024$ & $20.53\pm0.28$ & 7.97 & 9.29 \\
        \midrule
        \multicolumn{6}{@{}l}{\textit{Pixelization}} \\
        T4   & $48\times48$  & $384\times384$  & $5.14\pm0.08$ & 5.25 & 6.43 \\
        T4   & $64\times64$  & $512\times512$  & $8.44\pm0.02$ & 5.60 & 6.43 \\
        T4   & $128\times128$ & $1024\times1024$ & $41.53\pm0.13$ & 7.97 & 10.30 \\
        A100 & $128\times128$ & $1024\times1024$ & $5.32\pm0.01$ & 7.97 & 9.22 \\
        \bottomrule
    \end{tabular}
    \captionof{table}{PAPify with SD3.5 Medium at block width $p=8$. Peak allocated and reserved GPU memory are in GiB.}
    \label{tab:supp_costs}
\end{center}

\paragraph{Timing.}
The generation profiling protocol covers both stages, from the text prompt to the final pixel art. The archived T4 generation runs at $48\times48$ and $64\times64$ have an unspecified timing scope. Pixelization time covers palette extraction and refinement of an existing image into pixel art.

\subsection{PAPify with SDXL}

We test PAPify with SDXL on a Tesla T4. After one warm-up run for each grid size, we generate pixel art from ten prompts, one at a time. Table~\ref{tab:supp_sdxl_costs} reports the average generation time, its standard deviation, and peak GPU memory for runs with $8\times8$ blocks. The preference evaluation settings are listed in Tables~\ref{tab:supp_sdxl_config} and~\ref{tab:supp_papify_sdxl}.

\begin{center}
    \centering
    \small
    \begin{tabular}{@{}lrrrrr@{}}
        \toprule
        Grid & RGB canvas & Colors & Time (s) & Alloc. & Reserv. \\
        \midrule
        $48\times48$  & $384\times384$  & 10 & $33.94\pm0.47$ & 4.97 & 5.21 \\
        $64\times64$  & $512\times512$  & 16 & $39.05\pm0.28$ & 4.99 & 5.24 \\
        $128\times128$ & $1024\times1024$ & 32 & $85.37\pm0.19$ & 5.24 & 5.61 \\
        \bottomrule
    \end{tabular}
    \captionof{table}{PAPify with SDXL on a Tesla T4. Peak allocated and reserved GPU memory are in GiB.}
    \label{tab:supp_sdxl_costs}
\end{center}

\subsection{Baselines}

Table~\ref{tab:supp_baseline_costs} reports the average time per image. We time local pixelization on a 2.00\,GHz Intel Xeon CPU, from loading a $512\times512$ PNG to saving the result. For each API, we time 100 requests per task and resolution, from sending the HTTP request through parsing the response.

\begin{center}
    \centering
    \small
    \begin{tabular}{@{}lrrr@{}}
        \toprule
        Method & $48\times48$ & $64\times64$ & $128\times128$ \\
        \midrule
        \multicolumn{4}{@{}l}{\textit{Local pixelization on CPU}} \\
        Direct PAP & 0.05 & 0.11 & 0.57 \\
        AOP (core) & 0.26 & 0.26 & 0.56 \\
        APIP (reimpl.) & 0.35 & 0.70 & 2.72 \\
        MYOS (TestPro) & 0.58 & 0.86 & 3.87 \\
        PIA & 1.69 & 2.63 & 13.69 \\
        \midrule
        \multicolumn{4}{@{}l}{\textit{API generation}} \\
        FLUX & 4.59 & 4.62 & 4.62 \\
        GPTImage & 14.73 & 15.10 & 15.83 \\
        \midrule
        \multicolumn{4}{@{}l}{\textit{API pixelization}} \\
        FLUX & 4.58 & 6.72 & 6.51 \\
        GPTImage & 17.42 & 21.70 & 22.16 \\
        \bottomrule
    \end{tabular}
    \captionof{table}{Mean seconds per image for local and API baselines. Local means are over ten inputs per configuration.}
    \label{tab:supp_baseline_costs}
\end{center}

\subsection{Runtime and GPU Memory Comparison}
\label{sec:supp_efficiency_comparison}

\paragraph{Runtime.}
At $64\times64$, PAPify with SDXL averages 39.05\,s per image on a Tesla T4 (Table~\ref{tab:supp_sdxl_costs}).
\mbox{SD-$\pi$XL} uses the same grid by default and reports 1.5--2.5\,h per image on an RTX 4090 for 6,000--10,000 optimization epochs~\cite[Supplement, Sec.~1.2]{binninger2024sdpixl}. The lower end is 1.5\,h, or 5,400\,s. Dividing by 39.05\,s gives a ratio of $138.3$, rounded to about $140\times$ in the main paper's introduction.
VectorFusion reports 25\,min per SVG on an RTX 2080 Ti at batch size 1 for its SD + LIVE + SDS variant~\cite[Supplement, Sec.~I.3]{jain2023vectorfusion}.

\paragraph{GPU memory.}
In the same PAPify setup, with FP16 and CPU offloading, PyTorch peaks at 4.99\,GiB allocated and 5.24\,GiB reserved (Table~\ref{tab:supp_sdxl_costs}).
VectorFusion specifies a minimum GPU capacity of 10\,GB at batch size 1~\cite[Supplement, Sec.~I.3]{jain2023vectorfusion}. PAPify's 5.24\,GiB of peak reserved memory is about 5.63\,GB, so it corresponds to $5.63/10\approx0.56$, or roughly half of that capacity.
The \href{https://github.com/AlexandreBinninger/SD-piXL\#memory-requirements}{\mbox{SD-$\pi$XL} implementation guide} recommends a 24\,GB GPU. Comparing peak memory across methods would require measurements under a common protocol.

%% file: sec/supp/E_analyses.tex
\clearpage

\section{Additional Analyses}
\label{sec:supp_analyses}
We present additional analyses of PAPify and illustrate our findings with qualitative examples.

\subsection{Refinement Behavior and Design Choices}

\paragraph{Effect of the final projection.}
Even with the grid size and color budget specified in the prompt, the raw SD3.5 Medium output needs visible corrections to fit the target grid and palette (Fig.~\ref{fig:supp_final_pap}). PAPify's output already follows these constraints closely before the final projection. Across 50 images per grid, the mean squared error (MSE) between the output and its PAP projection drops by a factor of 29 to 108 after refinement. The intermediate predictions in Fig.~\ref{fig:supp_inference_process} show how pixel-art structure emerges over the denoising steps. The final projection enforces the grid and palette exactly.

\begin{center}
    \includegraphics[width=0.9\textwidth]{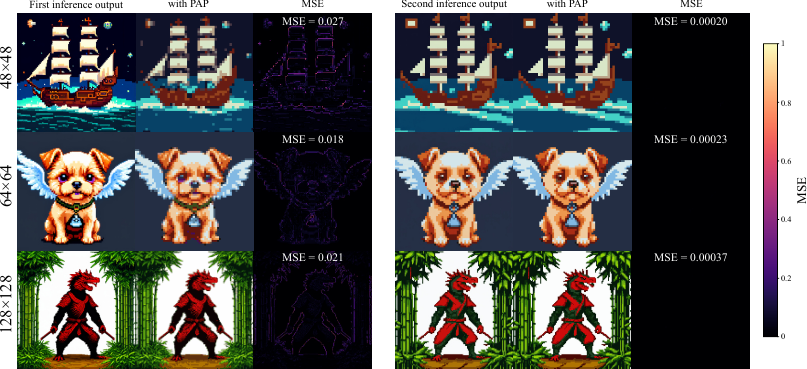}
    \captionof{figure}{\textbf{Correction induced by PAP before and after refinement.} Each stage shows the decoded output, its projection onto the target grid and palette, and the squared RGB error map. Projection visibly changes the first-stage outputs, especially along fine contours. After refinement, the decoded and projected images are visually close.}
    \label{fig:supp_final_pap}
\end{center}

\paragraph{Pixelization with PAPify.}
Figure~\ref{fig:supp_inference_process} shows PAPify pixelizing a photograph. The text conditioning consists only of the scene description ``a cute cat in the street.'' During denoising, the cat's outline becomes clearer and the street is rendered in larger regions of color. The pixel-art style is already visible before the final projection.

\begin{center}
    \begin{minipage}{\textwidth}
        \centering
        \includegraphics[width=\linewidth]{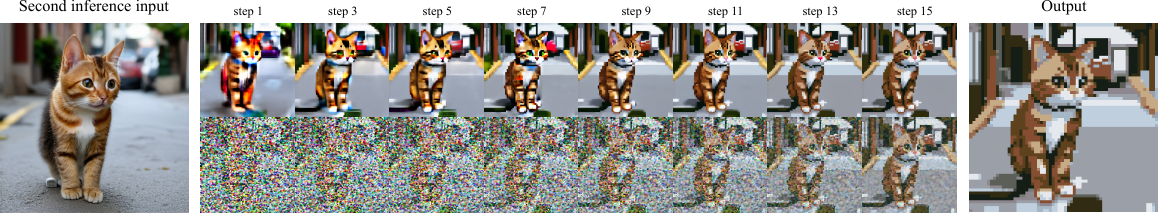}
        \captionof{figure}{\textbf{Pixelization of a photograph with PAPify.} At selected denoising steps, we show clean predictions (top) and noisy states (bottom), with the input photograph on the left and the final pixel art on the right. The prompt is ``a cute cat in the street.''}
        \label{fig:supp_inference_process}
    \end{minipage}
\end{center}

\clearpage

\paragraph{Flow shift.}
Figure~\ref{fig:supp_flow_shift} compares flow shifts of 3 and 12 using the same reference images. At 3, PAPify's outputs often resemble a direct PAP projection of the reference, with more local detail and blurry contours. At 12, PAPify can reinterpret the scene by redrawing objects and reorganizing details on the target grid. The resulting silhouettes are clearer and the color regions larger, bringing the output closer to pixel-art conventions (\mainpapersection{sec:pixel_flow_shift}{4.3}).

\begin{center}
    \includegraphics[width=\textwidth]{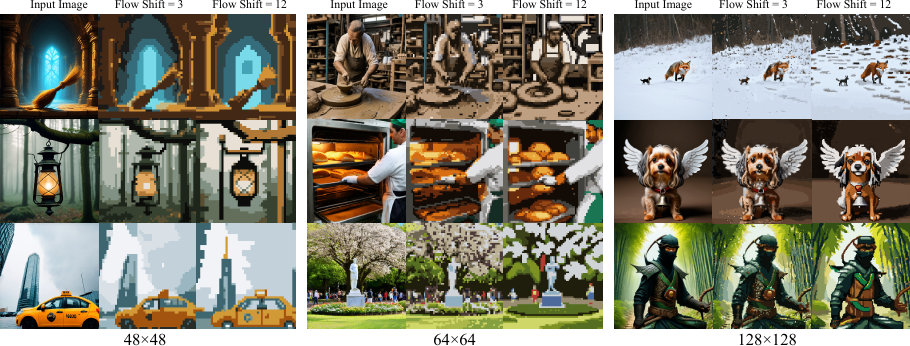}
    \captionof{figure}{\textbf{Flow shift during refinement.} The three groups show target grids of $48\times48$, $64\times64$, and $128\times128$ from left to right. Within each triplet, the reference is followed by the outputs at flow shifts of 3 and 12. The higher flow shift induces larger changes in contours and detail.}
    \label{fig:supp_flow_shift}
\end{center}

\paragraph{Canvas size and block width.}
Figure~\ref{fig:supp_resolution_tradeoff} compares two settings at the same $128\times128$ output resolution, with a $512\times512$ canvas for $p=4$ and a $1024\times1024$ canvas for $p=8$. With $p=4$, small features tend to merge into larger color regions. With $p=8$, finer detail remains visible down to individual output pixels. This is consistent with the VAE reconstruction artifacts observed for smaller blocks (\mainpapersection{sec:flowshift_analysis}{4.3.1}).

\begin{center}
    \includegraphics[width=\textwidth]{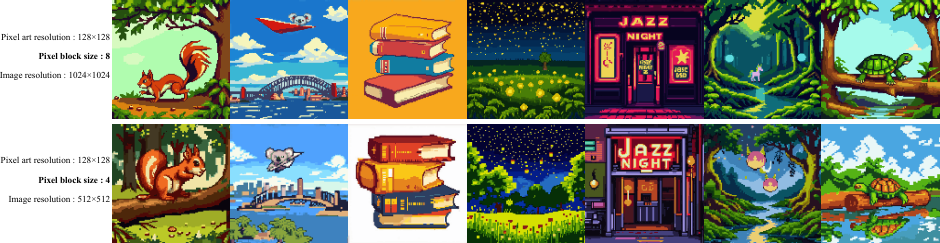}
    \captionof{figure}{\textbf{Canvas and block width at a fixed $128\times128$ grid.} Both rows use the same prompts. The $1024\times1024$ canvas with $p=8$ (top) yields finer structures and cleaner boundaries in these examples than the $512\times512$ canvas with $p=4$ (bottom).}
    \label{fig:supp_resolution_tradeoff}
\end{center}

\clearpage

\subsection{Comparison with Prompted Generators}

\paragraph{Comparison with SD3.5 Medium.}
Figure~\ref{fig:supp_sd35_comparison} illustrates how PAPify improves the pixel-art output of SD3.5 Medium. The qualitative comparison shows the prompted output, its direct PAP projection, and PAPify's output with the same backbone. Under the same target grid and color budget as direct PAP, PAPify's refinement produces clearer silhouettes and more coherent color regions.

\begin{center}
    \includegraphics[width=\textwidth]{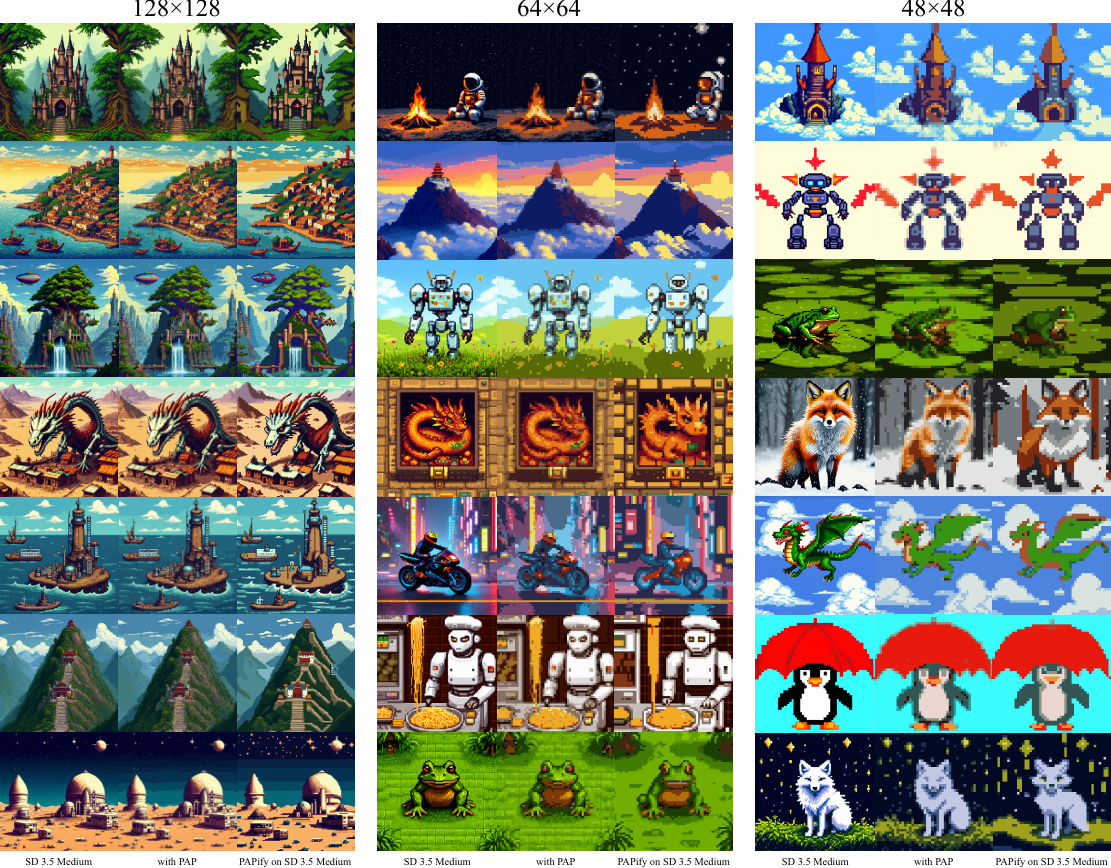}
    \captionof{figure}{\textbf{Stable Diffusion 3.5 Medium, direct PAP, and PAPify.} Each triplet shows the prompted baseline, its direct projection, and PAPify from left to right. The groups correspond to $128\times128$, $64\times64$, and $48\times48$ target grids.}
    \label{fig:supp_sd35_comparison}
\end{center}

\clearpage

\paragraph{Grid size and color budget in prompted outputs.}
We give FLUX and GPTImage the same scene descriptions with different target grid sizes and color budgets. The raw outputs have a similar pixel-art appearance across requests, with little change in apparent block size (Fig.~\ref{fig:supp_requested_resolution}). Counting distinct RGB colors across the benchmark, we find that all 300 generation outputs and all 300 pixelization outputs from each model exceed the requested color budget. Applying PAP enforces both constraints, with larger blocks at $48\times48$ and finer detail at $128\times128$.

\begin{center}
    \includegraphics[width=0.82\textwidth]{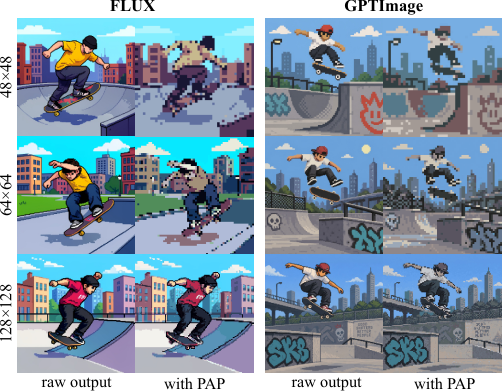}
    \captionof{figure}{\textbf{Prompted grid size in FLUX and GPTImage.} The rows show requests for $48\times48$, $64\times64$, and $128\times128$ grids. For each model and grid, the left image is the raw output and the right image is its PAP projection.}
    \label{fig:supp_requested_resolution}
\end{center}

%% file: sec/supp/F_results_limits.tex
\section{Limitations}
\label{sec:supp_limits}

The initial PAP projection can erase small objects or merge regions when the grid is coarse or the palette offers little contrast. 
For example, Merlot has five colors, the fewest among the fixed palettes in Fig.~\ref{fig:supp_merlot_limit}. With this palette, PAP maps most of the scene to one color, erasing the boundaries between the sky, water, and shore. Refinement may struggle to recover the lost structure.

\par\smallskip\noindent
\begin{minipage}{\textwidth}
    \centering
    \includegraphics[width=\linewidth]{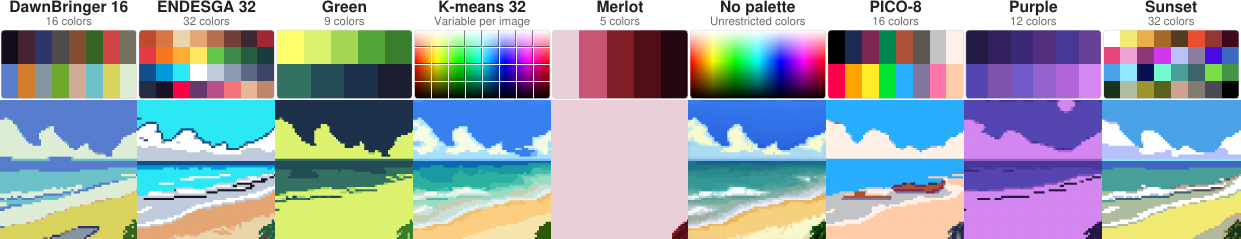}
    \captionsetup{hypcap=false,skip=5pt}
    \captionof{figure}{\textbf{Effect of palette choice.} Each column shows the same scene with a different palette setting.}
    \label{fig:supp_merlot_limit}
\end{minipage}

\par\smallskip